%% file: arxiv_main.tex
\pdfoutput=1

\documentclass[11pt]{article}

 \usepackage[preprint]{neurips_2025}
\usepackage{selectp}
\usepackage[utf8]{inputenc} 
\usepackage[T1]{fontenc}    
\usepackage[hidelinks, colorlinks=true,allcolors=teal]{hyperref}
\usepackage{url}            
\usepackage{booktabs}       
\usepackage{amsfonts}       
\usepackage{nicefrac}       
\usepackage{microtype}      
\usepackage{xcolor}         

\usepackage{times}
\usepackage{latexsym}

\usepackage{inconsolata}

\input{math_commands.tex}

\usepackage{algorithmic}
\usepackage{graphicx}
\usepackage{booktabs} 

\usepackage{amsmath}
\usepackage{amssymb}
\usepackage{mathtools}
\usepackage{amsthm}
\usepackage{algorithm}
\usepackage[capitalize,noabbrev]{cleveref}
\usepackage{makecell}
\usepackage[acronym,nohypertypes={acronym,notation}]{glossaries}
\glsdisablehyper
\input{glossary.tex}
\usepackage{wrapfig}
\usepackage[subtle]{savetrees}
\usepackage[compact]{titlesec}
\titlespacing\subsection{0pt}{8pt plus 4pt minus 2pt}{0pt plus 2pt minus 2pt}
\usepackage{subcaption}
\usepackage{multirow}
\usepackage{hhline}
\usepackage[inline,shortlabels]{enumitem}
\usepackage{listings}
\usepackage{placeins}
\setlist{nolistsep}
\usepackage[obeyspaces]{xurl}
\usepackage[textsize=tiny]{todonotes}
\expandafter\def\expandafter\normalsize\expandafter{%
    \normalsize%
    \setlength\abovedisplayskip{0pt}%
    \setlength\belowdisplayskip{8pt}%
    \setlength\abovedisplayshortskip{-8pt}%
    \setlength\belowdisplayshortskip{2pt}%
}

\theoremstyle{plain}

\theoremstyle{definition}

\theoremstyle{remark}

\title{SD$^2$: Self-Distilled Sparse Drafters}

\author{
    Mike Lasby$^{1,2,\dagger}$, Nish Sinnadurai$^{1}$, Valavan Manohararajah$^{1}$,\\ \textbf{Sean Lie$^{1}$, Yani Ioannou$^{2}$, Vithursan Thangarasa$^{1}$}\\
    $^{1}$Cerebras Systems Inc., $^{2}$Schulich School of Engineering, University of Calgary
}

\begin{document}
\maketitle

\newcommand\blfootnote[1]{%
  \begingroup
  \renewcommand\thefootnote{}\footnote{#1}%
  \addtocounter{footnote}{-1}%
  \endgroup
}   
\blfootnote{Correspondence to \href{mailto:mklasby@ucalgary.ca}{mklasby@ucalgary.ca}}
\blfootnote{$\dagger$ Work completed while on internship at Cerebras}

\begin{abstract}
Speculative decoding is a powerful technique for reducing the latency of Large Language Models (LLMs), offering a fault-tolerant framework that enables the use of highly compressed draft models. In this work, we introduce Self-Distilled Sparse Drafters (SD$^2$), a novel methodology that leverages self-data distillation and fine-grained weight sparsity to produce highly efficient and well-aligned draft models. SD$^2$ systematically enhances draft token acceptance rates while significantly reducing Multiply-Accumulate operations (MACs), even in the Universal Assisted Generation (UAG) setting, where draft and target models originate from different model families. On a Llama-3.1-70B target model, SD$^2$ provides a 1.59$\times$ higher Mean Accepted Length (MAL) compared to layer-pruned draft models and reduces MACs by over 43.87\% with a 8.36\% reduction in MAL compared to a dense draft models. Our 1.5B and 3B unstructured sparse drafters outperform both dense and layer-pruned models in terms of end-to-end latency improvements; highlighting the potential of sparsity-aware fine-tuning and compression strategies to improve LLM inference efficiency while maintaining alignment with target models.
\end{abstract}

\glsresetall
\section{Introduction}\label{sec:intro}
\glspl{llm} have proven to have high utility in a wide variety of contexts. However, the causal dependency between preceding and subsequent tokens results in approximately 10$\times$ higher latency for sequence generation compared to processing an equivalent length sequence in parallel~\citep{liu_deja_2023}. The high computational cost of \glspl{llm} has motivated significant research into methods which improve their efficiency, including: quantization~\citep{gholami_survey_2021,kurtic_give_2024}, 
pruning~\citep{ma_llm-pruner_2023,gromov_unreasonable_2024},
weight sparsity~\citep{frantar_sparsegpt_2023,sun_simple_2023,yin_outlier_2023}, 
activation sparsity~\citep{mirzadeh_relu_2023},
KV-cache compression~\citep{zhang_q-hitter_2024}, 
distillation~\citep{kim_sequence-level_2016, hsieh_distilling_2023}, 
and matrix decomposition~\citep{hu_lora_2021,liu_dora_2024}. However, these methods typically trade improved efficiency for decreased model quality.~\citep{yin_junk_2023,jaiswal_compressing_2023}. 

In contrast, speculative decoding~\citep{stern_blockwise_2018,leviathan_fast_2023,chen_accelerating_2023} offers a unique framework to accelerate token generation \textit{without} sacrificing accuracy. In speculative decoding, a smaller \textit{draft} model is utilized to auto-regressively generate a sequence of \textit{draft tokens} which are verified in parallel by a \textit{target} model. For speculative decoding to be effective, an efficient draft model which is closely aligned with the target model is required. How best to select and/or train a draft model has been the focus of several recent works, see~\citet{xia_unlocking_2024} and \cref{sec:related-work} for more details.

\begin{figure*}[th]
     \centering
     \includegraphics[width=\textwidth]{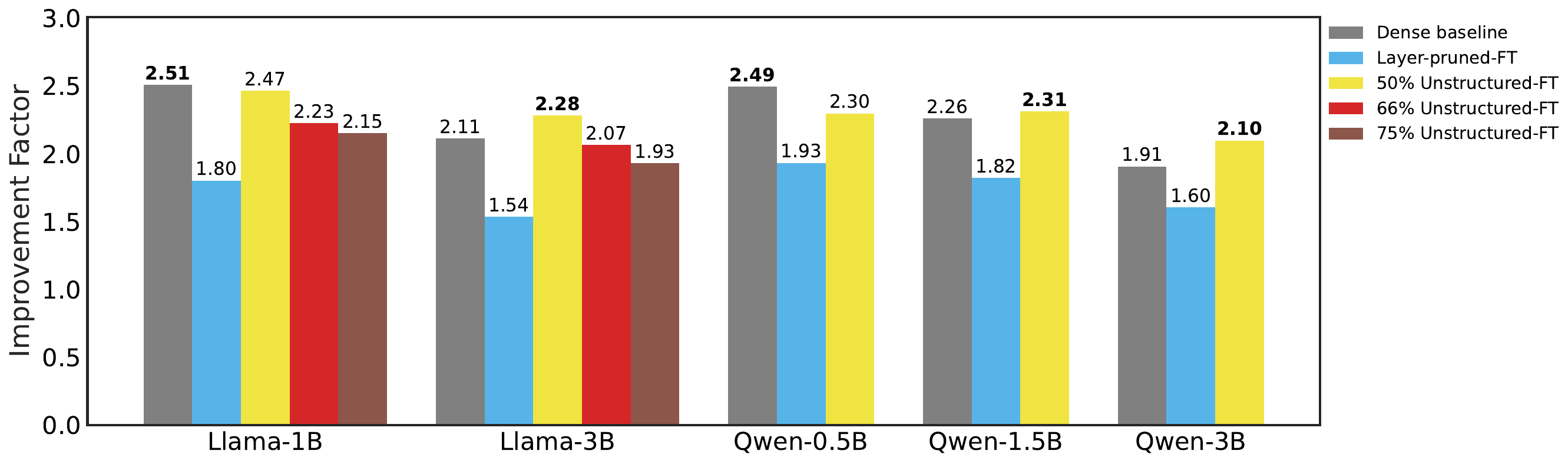}
     \caption{\textbf{Improvement factor of dense, layer-pruned, and \glstext{sdsd} unstructured} Llama and Qwen models drafting for Llama-3.1-70B-Instruct and Qwen-2.5-72B-Instruct, respectively. \glstext{sdsd} drafters outperform layer-pruned draft models and dense drafters in the 1.5B and 3B model size categories.} 
     \label{fig:nm_vllm_combined}
 \end{figure*}

\newlength{\figwidth}
\newlength{\figheight}
\setlength{\figwidth}{0.33\textwidth}
\setlength{\figheight}{0.22\textwidth}

\begin{figure*}[t]
    \begin{subfigure}{0.49\textwidth}
        \centering
        \includegraphics[width=1.0\textwidth]{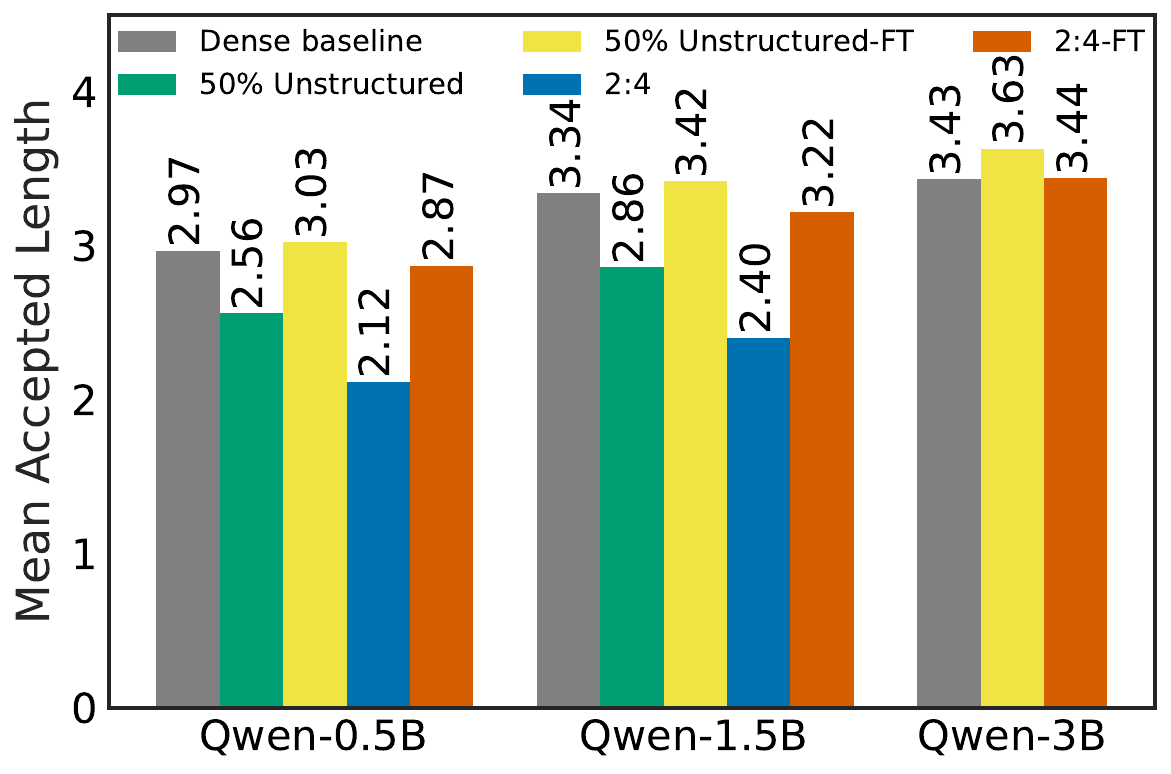}
        \vspace{-0.659em}
        \caption{\textbf{\glstext{uag} Qwen drafting for Llama}}
        \label{fig:uag-spec}
    \end{subfigure}
    \begin{subfigure}{0.49\textwidth}
         \centering
        \includegraphics[width=1.0\textwidth]{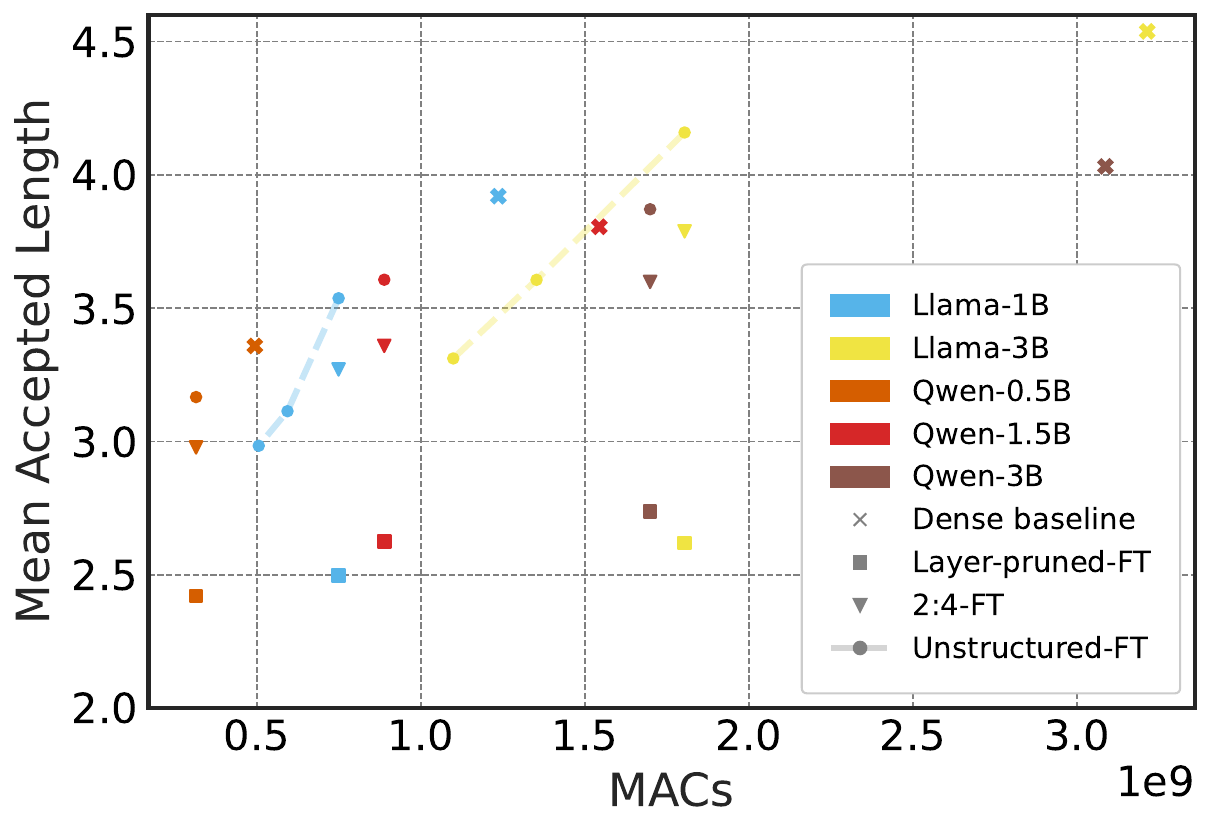}
        \caption{\textbf{MACs analysis}}
        \label{fig:macs}
    \end{subfigure}
    \caption{\textbf{(\subref{fig:uag-spec}): SpecBench \glstext{mal} for \glstext{sdsd} Qwen-2.5 models} drafting for Llama-3.1-70B-Instruct in the \glstext{uag} setting. These results illustrate the benefits of \glstext{sdsd} for aligning draft models \textit{even across different model families}. \glstext{sdsd} Qwen drafters achieve a \textbf{higher} \glstext{mal} than their dense counterparts. \textbf{(\subref{fig:macs}): \glstext{mal} vs. \glsentryplural{mac} for layer-pruned and sparse draft models.} Particularly notable are the Qwen-2.5 unstructured sparse drafters which approach iso-\glstext{mac} performance compared to the dense models.} 
\end{figure*}

Fine-grained sparsity, such as unstructured or 2:4~\citep{mishra_accelerating_2021} sparsity, for compressing draft models has yet to be examined. \glspl{snn} have significantly reduced connectivity between neurons in adjacent layers compared to dense networks. In unstructured sparsity the active parameters are distributed in an irregular, non-uniform manner throughout the weight matrices which can be challenging to accelerate. To address this, 2:4 sparsity was introduced which offers a hardware-friendly structure in which exactly two out of every four contiguous weights are active. However, the additional constraints of 2:4 sparsity result in lower accuracy on downstream tasks compared to unstructured.

Speculative decoding is a uniquely well-suited setting for using sparse draft models since any errors produced during drafting can be gracefully recovered from during the verification step. Other compression strategies such as quantization succeed in reducing memory overhead, but may not reduce latency~\citep{kurtic_give_2024}. Performing operations with parameters quantized to low bit-width data types requires dequantization to a hardware-native data type, typically 8- or 16-bit types for presently available accelerators. As a result, quantization may lead to increased latency in many circumstances, particularly for relatively small draft models which may not be memory-bound. In contrast, \glspl{snn} reduce the total amount of \glspl{mac} required per forward pass which, in an ideal setting, would correspond directly with latency improvements. In practice, accessing non-contiguous sparse parameters and storing the associated sparse structure metadata can lead to significant overhead on \glspl{gpu}~\citep{hooker_hardware}. Achieving real-world acceleration for fine-grained \glspl{snn} requires specialized kernels~\citep{neural_magic_neuralmagicdeepsparse_2021,schultheis_towards_2023,lasby_dynamic_2023,frantar_marlin_2024} and hardware such as Cerebras' wafer-scale engine~\citep{thangarasa_sparse_ift_2024,cerebras_training,lie_cerebras_2023,agarwalla_enabling_2024}.

Given the practical limitations of fine-grained sparsity, \textit{structured pruning} --- the removal of layers, neurons, or other substructures --- of \glspl{llm} to obtain performant draft models is a more hardware-friendly alternative. In particular, pruning entire transformer \textit{blocks} from the decoder of \glspl{llm} provides low latency models which retain their quality under moderate compression ratios~\citep{liu_deja_2023,gromov_unreasonable_2024,men_shortgpt_2024,kim_shortened_2024,sun_transformer_2024}. A natural extension of these works is to leverage layer-pruning techniques to obtain draft models as examined by ~\citet{thangarasa2024selfdatadistillationrecoveringquality}. 
However, it is not clear a priori if the model quality degradation of layer-pruned models can be overcome by their improved latency for use as draft models in the speculative decoding setting. 

In this work, we propose \glsreset{sdsd}\gls{sdsd} --- a novel methodology for obtaining efficient, well-aligned draft models --- by utilizing \textit{self-data distillation fine-tuning}~\citep{yang_self-distillation_2024,thangarasa2024selfdatadistillationrecoveringquality} and weight sparsity. Specifically, we make the following contributions: 

\begin{itemize}
    \item We introduce \gls{sdsd}, a novel methodology for obtaining fine-grained sparse draft models;
    \item We demonstrate the superiority of fine-grained sparsity for accelerating speculative decoding and downstream evaluation tasks compared with layer-pruned models; 
    \item We showcase the effectiveness of self-data distillation fine-tuning for model alignment, even when aligning with a different model family with \gls{uag};
    \item When paired with optimized sparse representations, we find that the end-to-end acceleration of speculative decoding with fine-grained sparse draft models is comparable to and in some cases exceeds that of dense draft models, particularly with larger draft model sizes.
\end{itemize}

\section{Method}\label{sec:method}
In this section, we provide preliminaries for speculative decoding and introduce the components of \gls{sdsd}, consisting of self-data distillation, one-shot fine-grained pruning, and sparse fine-tuning. We also present our layer pruning method which serves as a baseline to compare to our sparse drafters.  

\subsection{Speculative decoding}
The original motivation for speculative decoding stems from the observation that a wide degree of variance exists in the difficulty of generating tokens. For instance, completions may require copy-pasting portions of the input or including tokens which do not contribute to the semantic content. However, in typical auto-regressive sampling each token requires precisely the same amount of computation to generate regardless of apparent difficulty. In speculative decoding, draft tokens are produced by auto-regressively sampling from a smaller draft model, $M_d$, to produce candidate completions which are verified in parallel by the target model, $M_t$. The \textit{draft-then-verify} procedure is repeated for multiple rounds until the generation outputs the end-of-sequence token or another stopping condition occurs -- such as reaching a maximum number of generated tokens.

Formally, given an input sequence $\{x_1, x_2, ... , x_n\}$, the draft model calculates probability distributions for each token in the completion conditioned on the input sequence and any preceding output tokens: $p_{n+j+1} = M_d(\{x_1, ... , x_n,... \tilde{x}_{n+j}\}) \forall{j} \in \{1, ..., k\}$. From these distributions draft tokens are sampled $\tilde{x}_{n+j}\sim p_{n+j}$ to generate a partial completion $\{\tilde{x}_{n+1}, ...,\tilde{x}_{n+k}\}$ of $k$ draft tokens. In the verification stage, the target model $M_t$ computes the output probabilities for each draft token plus one additional token based on its own output distribution: 
$q_{j+1} = M_t(x_{\leq{n}}, \tilde{x}_{\leq{j}}) \forall{j} \in \{n,...,n+k+1\}$.
For each draft token, $\tilde{x_j}$, the token is accepted if a verification condition based on the draft and target model probabilities is satisfied.
A variety of sampling and verification schemes have been considered in prior work~\citep{stern_blockwise_2018,leviathan_fast_2023,chen_accelerating_2023}.
In our experiments, we use greedy sampling with a strict top-1 verification criterion which guarantees that generated text matches the output of the original target model precisely:

\begin{equation}
\tilde{x}_{n+j} = \argmax \; p_j \quad \forall{j} \in \{1,...,k\},
\end{equation}

\begin{equation}
x_{n+j} = \begin{cases}
    \tilde{x}_{n+j} & \text{if $\tilde{x}_{n+j} = \argmax  \; q_{n+j}$} \\
    \argmax \; q_{n+j} & \text{otherwise}.
\end{cases}
\end{equation}

We model the expected \textit{improvement factor} to the overall latency of speculative decoding versus sampling from the target model directly as follows: 

\begin{equation}\label{eq:improvement-factor}
    \text{Improvement Factor} = \frac{\text{MAL}}{kc+1},
\end{equation}

where \gls{mal} is the average number of tokens accepted per round\footnote{\gls{mal} includes accepted draft tokens and one additional token from the target model per round.}, $k$ is the number of draft tokens speculated per round, and $c$ is the cost factor representing the ratio of draft versus target model efficiency. For our practical improvement factor calculations, $c$ represents the wall clock latency ratio, i.e., $c = \frac{\text{timeit}(M_d(x))}{\text{timeit}(M_t(x))}$. For our theoretical improvement factor calculations, we substitute \glspl{mac} in lieu of latencies.

\begin{algorithm}[tb]
\caption{Self-data distillation for speculative decoding}\label{alg:self-data}
\begin{algorithmic}[1]
\STATE {\bfseries Input:}  Pretrained target model $M_t$, fine-tuning dataset $D_{f}$ with prompts $\mathcal{X}$, context $\mathcal{C}$, and labels $\mathcal{Y}$
\STATE Initialize $D_{self} = \emptyset$
\FOR{$\mathbf{X_i}, \mathbf{Y_i}, \mathbf{C_i} \in D_f$}
    \STATE $\mathbf{X'_i} \gets \mathbin\Vert \mathbf{C_i} \mathbin\Vert\mathbf{X_i} \mathbin\Vert \mathbf{Y_i}$
    \small \COMMENT{Combine into new prompt} \normalsize
    \STATE $\mathbf{\Tilde{Y_i}} \sim M_t(\mathbf{X'_i})$ 
    \small \COMMENT{Generate new label} \normalsize
    \STATE $D_{self} \gets (\mathbf{X_i}, \mathbf{\Tilde{Y_i}})$ \small \COMMENT{Accept $\mathbf{\Tilde{Y_i}}$ w/o verification} \normalsize
\ENDFOR
\STATE {\bfseries Output:} $D_{self}$
\end{algorithmic}
\end{algorithm}

\subsection{Self-data distillation}

Self-data distillation consists of generating a dataset, $D_{self}$, whose outputs are generated by a model of interest. We curate fine-tuning datasets following ~\citet{yang_self-distillation_2024, thangarasa2024selfdatadistillationrecoveringquality}. The distilled labels are generated by the target model, $M_t$ based on the input sequences, ground truth labels, and task-specific context of one or more supervised fine-tuning datasets $D_{f}$. Specifically, given a task specific context $\mathbf{C^t}$, original input sequence $\mathbf{X^t}$, and original ground truth label $\mathbf{Y^t}$. We combine these components into a new input sequence $\mathbf{X'}$

\begin{equation}
    \mathbf{\tilde{Y}} = M_t(\mathbf{X'}) \quad \text{where} \; \mathbf{X}' = \mathbf{C^t}\mathbin\Vert\mathbf{X^t}\mathbin\Vert \mathbf{Y^t}.
\end{equation}

In the original formulation of self-data distillation, \citet{yang_self-distillation_2024} only extract the distilled label $\mathbf{\tilde{Y}}$ if and only if it aligns with the original ground truth label, ${\mathbf{Y^t}}$. For instance, only accepting the distilled label if it is mathematically equivalent to the ground truth label. However, for tasks that do not yield a closed-form solution, it can be challenging to assess the validity of the distilled label. 

The speculative decoding setting gracefully eliminates the need to consider the distilled label verification process. Since our fundamental goal is aligning the draft with the target model, the correctness of the distilled label is irrelevant. We accept the distilled labels even if they contain easily identifiable errors; the distilled output, correct or otherwise, is already aligned with the target model output distribution. See \cref{alg:self-data} for a summary of the self-data distillation process.

\subsection{Fine-grained pruning}
To obtain our fine-grained sparse draft models, $M_d$, we use SparseGPT~\citep{frantar_sparsegpt_2023}. However, we emphasize that our method is compatible with any pruning algorithm. In addition to the original SparseGPT hyperparameter settings, we experiment with non-uniform layer-wise sparsity distributions such as \gls{owl}~\citep{yin_outlier_2023} and a novel distribution inspired by the angular cosine distance measure from ~\citet{gromov_unreasonable_2024}. While our experiments found that non-uniform layer-wise sparsity improves perplexity, little benefit was observed on downstream evaluations or speculative decoding. See \cref{sec:pruning-results} for a more thorough discussion of our findings. Ultimately, we use a uniform layer-wise sparsity distribution for all results presented below.

\subsection{Sparse fine-tuning}\label{sec:sparse-sft}
In typical supervised fine-tuning settings, a dense \gls{llm} $M$ with parameters $\theta$ is fine-tuned using a supervised dataset $D_{f}$ containing input sequences $\mathcal{X}$ and ground truth output sequences $\mathcal{Y}$. For each token, $y_j$, in the ground truth output sequences, $\mathbf{Y_i} \in \mathcal{Y}$, the model outputs a probability distribution over its vocabulary conditioned by the input sequences, $\mathbf{X_i} \in \mathcal{X}$, any output tokens preceding the current token position, and the model parameters: $Q({y}_{j} | \mathbf{X_i}, \{y_1, ..., y_{j-1}\}, \theta)$ where $Q(\cdot)$ represents the \gls{lm-head} logits normalized with the softmax function. The total loss per mini-batch is the average negative log-likelihood across all sequences and tokens:

\begin{equation}
    \mathcal{L} = - \frac{1}{N} \sum_{i=1}^N \frac{1}{S_i} \sum_{j=1}^{S_i-1} P(y_{j})\log(Q({y}_{j} | \mathbf{X_i}, y_{\leq j-1}, \theta)),
\end{equation}

where $P(y_{j})$ is the ground truth distribution, $N$ is the number of samples in the batch, and $S_i$ is the output sequence length. The model is fine-tuned by minimizing this loss using stochastic gradient descent. In the sparse fine-tuning setting, we simply replace $\theta$ with $\theta_{s} \subset \theta$, the set of non-zero parameters. In our approach, the sparse topology is fixed after one-shot pruning. Pruned parameters remain fixed at zero throughout fine-tuning.

For our experiments, we fine-tune our sparse models with a binary mask to initialize pruned parameters to zero and set their gradients to zero during backpropagation via backwards hook. The backwards hook ensures that gradients of pruned parameters do not contribute to the partial derivates of active parameters nor optimizer buffer states. See \cref{alg:sparse-ft} for more details. 

\newenvironment{myindent}
  {\list{}{\leftmargin=1em \itemindent=0em
    \topsep=0pt \partopsep=0pt \itemsep=0pt \parsep=0pt}%
    \item[]
  }
  {\endlist}
  
\begin{algorithm}[tb]
\caption{Sparse fine-tuning}\label{alg:sparse-ft}
\begin{algorithmic}
\STATE {\bfseries Input:} Pruned draft model $M_d'$ with trainable parameters $\theta$, pruned parameters $\theta_p$, self-data distilled dataset $D_{self} = \{(\mathbf{X}_i, \mathbf{\tilde{Y}}_i)\}_{i=1}^{T*N}$ with distilled outputs sequences of length $S_i$, optimizer $\mathcal{O}$ (e.g., AdamW), learning rate $\alpha$, number of iterations $T$, and batch size $N$.

\STATE {\bfseries Define} \texttt{GradientHook(}$\theta,\theta_p,\nabla_\theta\mathcal{L}_t$\texttt{)}:
    \begin{myindent}
    \vspace{-1.25em}
    \IF{$\theta\texttt{.backwards()}$}
        \FOR{$p_i, \frac{\partial \mathcal{L}_t}{\partial p_i} \in \{\theta, \nabla_\theta\mathcal{L}_t\}$}
            \IF{$p_i \in \theta_p$}
                \STATE $\frac{\partial \mathcal{L}_t}{\partial p_i} \gets 0$
            \ENDIF
        \ENDFOR
    \ENDIF
    \STATE \textbf{return} $\nabla_{\theta_s}\mathcal{L}_t$
    \end{myindent}
\STATE $\theta_{s} \gets \theta \notin \theta_p$ 
\small \COMMENT{Set of active parameters} \normalsize
\STATE $\theta\texttt{.register(GradientHook)}$
\FOR{$t = 1$ {\bfseries to} $T$}
    \STATE $(\mathbf{X}_i, \mathbf{\tilde{Y}}_i) \sim D_{self}$
    \small \COMMENT{Sample mini-batch} \normalsize
    \STATE $\mathcal{L}_t \gets 0$ 
    \small \COMMENT{Initialize mini-batch loss} \normalsize
    \FOR{$n=1$ {\bfseries to} $N$}
        \STATE ${L}_n \gets 0$ 
        \small \COMMENT{Initialize sequence loss} \normalsize
        \FOR{$j=1$ {\bfseries to} $S_{i}-1$}
            \STATE $L_n \mathrel{{+}{=}} P(\tilde{y}_j)\log Q(\tilde{y}_j| \mathbf{X_i}, \tilde{y}_{\leq j-1}, \theta_{s}) $
        \ENDFOR
    \STATE $\mathcal{L}_t \mathrel{{+}{=}} (L_n/(S_i-1))/N$
    \ENDFOR
    \STATE $\nabla_{\theta_{s}}\mathcal{L}_t = \mathcal{L}_t\texttt{.backwards()}$
    \small \COMMENT{Triggers grad hook}\normalsize
    \STATE $\theta_{s} \gets \mathcal{O}(\theta_{s}, \nabla_{\theta_{s}}\mathcal{L}_t, \alpha)$
\ENDFOR
\STATE {\bfseries Output:} $M_d'$
\end{algorithmic}
\end{algorithm}

\subsection{Layer pruning}
We obtain our layer-pruned baselines as follows. Consider a draft model $M_d$ with $N$ decoder blocks. Each decoder block consists of a multi-headed self-attention module followed by a feed-forward module. The decoder blocks are stacked sequentially such that the output of the preceding block is the input to the following block. The final block's output is typically used for downstream tasks such as language modelling or sequence classification. The goal of layer pruning is to identify and prune $n$ sequential decoder blocks such that the resultant quality degradation on relevant downstream tasks is minimized. 

Various approaches have been proposed for identifying the most important blocks in a transformer~\citep{samragh2023weightsubcloningdirectinitialization,men_shortgpt_2024}. Following \citet{thangarasa2024selfdatadistillationrecoveringquality}, we elect to use the angular cosine distance measure as proposed by ~\citet{gromov_unreasonable_2024}. This metric uses the angular cosine distance between the input and output of a group of $n$ sequential decoder blocks to define block importance. The group of blocks with the highest similarity between their input and output are considered to be more redundant than other group candidates and therefore can be pruned with the smallest impact on the overall model output. 
Formally, we define the angular cosine distance measure as follows:

\begin{equation}\label{eq:angular-dist}
d(x_{D_t}^{i}\text{,}\; x_{D_t}^{i+n}) = \frac{1}{\pi} \text{arccos} \left(\frac{x_{D_t}^i \cdot x_{D_t}^{i+n}}{\left \| x_{D_t}^i \right \|\left \| x_{D_t}^{i+n} \right \|}\right),
\end{equation}

where $x_{D_t}^{i}$ and $x_{D_t}^{i+n}$ are vectors representing the inputs to blocks $i$ and $i+n$, respectively, for the last token $t$ in each sequence across a representative calibration dataset $D$. The dot product of these vectors is normalized using the $L^2$ norm ($\left \| \cdot \right \|$) to facilitate comparisons between different groups of blocks. To identify the optimal $i$ to begin pruning, we simply identify the group of blocks with the smallest angular distance: 

\begin{equation}
i^*(n) = \argmin_i \; d(x_{D_t}^{i}\text{,}\; x_{D_t}^{i+n}),
\end{equation}

where $i^*(n)$ is the starting block index for the block group of length $n$ with the minimal angular cosine distance for a given calibration dataset subset. Once identified, decoder blocks $i^*$ to $i^*+(n-1)$ are pruned and the outputs of layer $i^*$ connect as input to block $i^*+n$ to obtain the pruned draft model $M_d'$.

\setlength{\figwidth}{0.33\textwidth}
\setlength{\figheight}{0.22\textwidth}

\begin{figure*}[t]
    \begin{subfigure}{1.0\textwidth}
        \centering
        \includegraphics[width=1.0\textwidth]{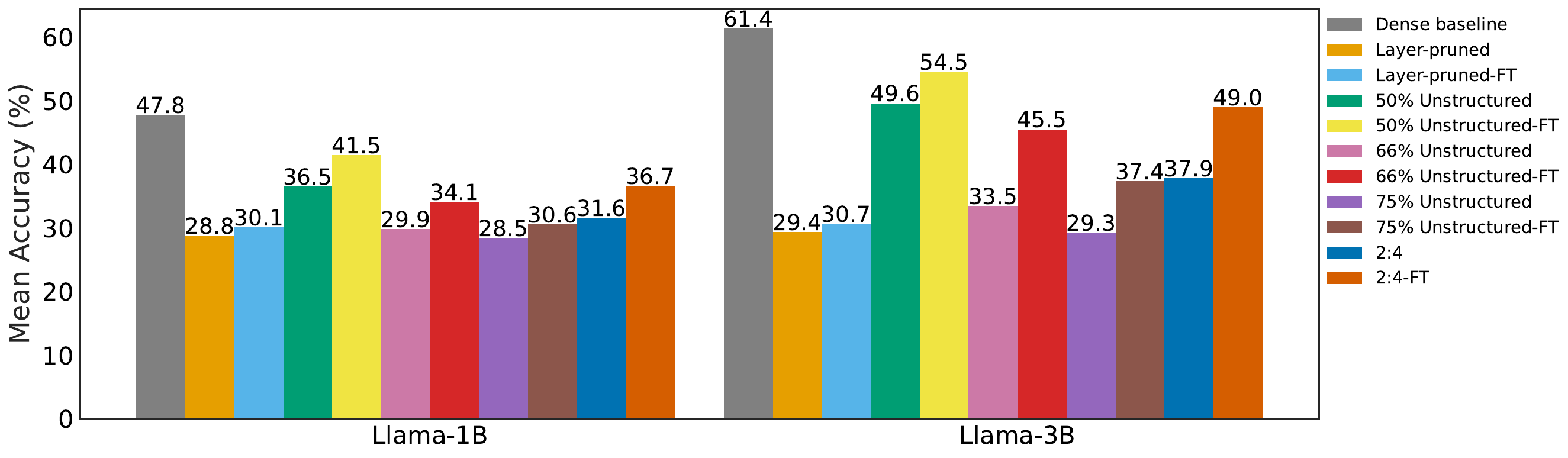}
        \caption{\textbf{Llama-3.2}}  
        \label{fig:llama-eval}
    \end{subfigure}
    \begin{subfigure}{1.0\textwidth}
         \centering
        \includegraphics[width=1.0\textwidth]{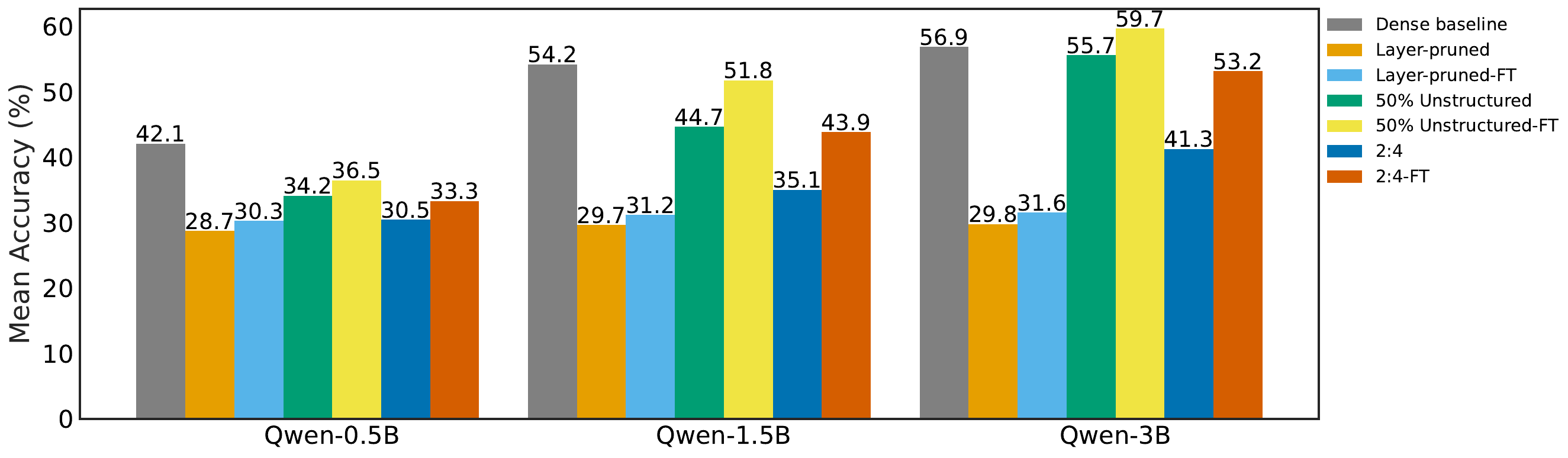}
        \caption{\textbf{Qwen-2.5}}  
        \label{fig:qwen-eval}
    \end{subfigure}
    \caption{\textbf{Average accuracy on OpenLLM Leaderboard V1} benchmarks for dense, layer-pruned, and sparse Llama-3.2 (\cref{fig:llama-eval}) and Qwen-2.5 (\cref{fig:qwen-eval}) draft models. Lighter and darker shades show results after one-shot pruning and \gls{sdsd} fine-tuning, respectively. Self-data distillation with sparse fine-tuning enables recovery of accuracy post-pruning. Notably, Qwen-2.5-3B-Instruct at 50\% unstructured sparsity \textit{exceeds the accuracy of the original dense model} after preparation with \gls{sdsd}} 
    \label{fig:evals}
\end{figure*}

\section{Results}\label{sec:results}
To empirically evaluate \gls{sdsd} and our primary hypothesis that sparse draft models can outperform dense or layer-pruned models, we evaluate a variety of draft model candidates across the Llama-3.2~\citep{llama3} and Qwen-2.5~\citep{qwen25} model families. For Llama, we evaluate drafting with Llama-3.2 1B-Instruct and 3B-Instruct for Llama-3.1-70B-Instruct. For Qwen-2.5, we evaluate 0.5B-Instruct, 1.5B-Instruct, 3B-Instruct drafting for Qwen-2.5-72B-Instruct. We also consider the \gls{uag} setting in which the Qwen draft models are aligned to and evaluated with a Llama-3.1-70B-Instruct target model. We prune and fine-tune fine-grained sparse draft models with both unstructured and 2:4 sparsity using a uniform layer-wise sparsity distribution and prune all decoder blocks, excluding the embedding and \gls{lm-head}. The sparsity levels reported in our results refer to the overall sparsity of the decoder. As baselines, we compare these candidates with 50\% layer-pruned\footnote{Specifically, 50\% of the decoder blocks are pruned.} and dense models. See \cref{sec:hyperparameters} for hyperparameters and implementation details.

\subsection{Evaluation}
We evaluate the draft models for downstream task accuracy on the OpenLLM Leaderboard V1 benchmarks~\citep{openllmleaderboard} and SpecBench~\citep{xia_unlocking_2024}. 

\subsubsection{OpenLLM leaderboard V1 benchmarks}
We evaluate our models using the default multi-shot settings on the OpenLLM Leaderboard V1 tasks using the EleutherAI evaluation harness~\citep{eval-harness}. These tasks include: 25-shot ARC-C~\citep{arc-c}, 5-shot strict exact match GSM8k~\citep{gsm8k}, 10-shot HellaSwag~\citep{hellaswag}, 5-shot MMLU~\citep{mmlu}, 5-shot Winogrande~\citep{winogrande}, and 0-shot multi-true (MC2) TruthfulQA~\citep{truthfulqa}. We report byte-length normalized accuracies for ARC-C and HellaSwag\footnote{Reported as the \texttt{acc\_norm} field in the EleutherAI evaluation harness outputs. See ~\citet{gao_multiple_2021} for details.}.

See \cref{fig:evals} for the mean accuracy for models pruned and fine-tuned using the \gls{sdsd} methodology. We find that one-shot pruned models suffer high degradation on the tasks evaluated; however, after fine-tuning the fine-grained sparse models recover much of their accuracy, particularly the 50\% unstructured and 2:4 sparse models. For the Qwen-2.5 1.5 and 3B models, we find that the unstructured 50\% models approach or \textit{exceed} the accuracy of the dense baseline. Notably, the 50\% layer-pruned models suffer high degradation, even after fine-tuning. See \cref{sec:eval-details} for more detailed results.

\setlength{\figwidth}{0.33\textwidth}
\setlength{\figheight}{0.22\textwidth}

\begin{figure*}[tb]
  \centering
  \begin{minipage}[b]{\figwidth}
    \centering
    \includegraphics[width=\figwidth]{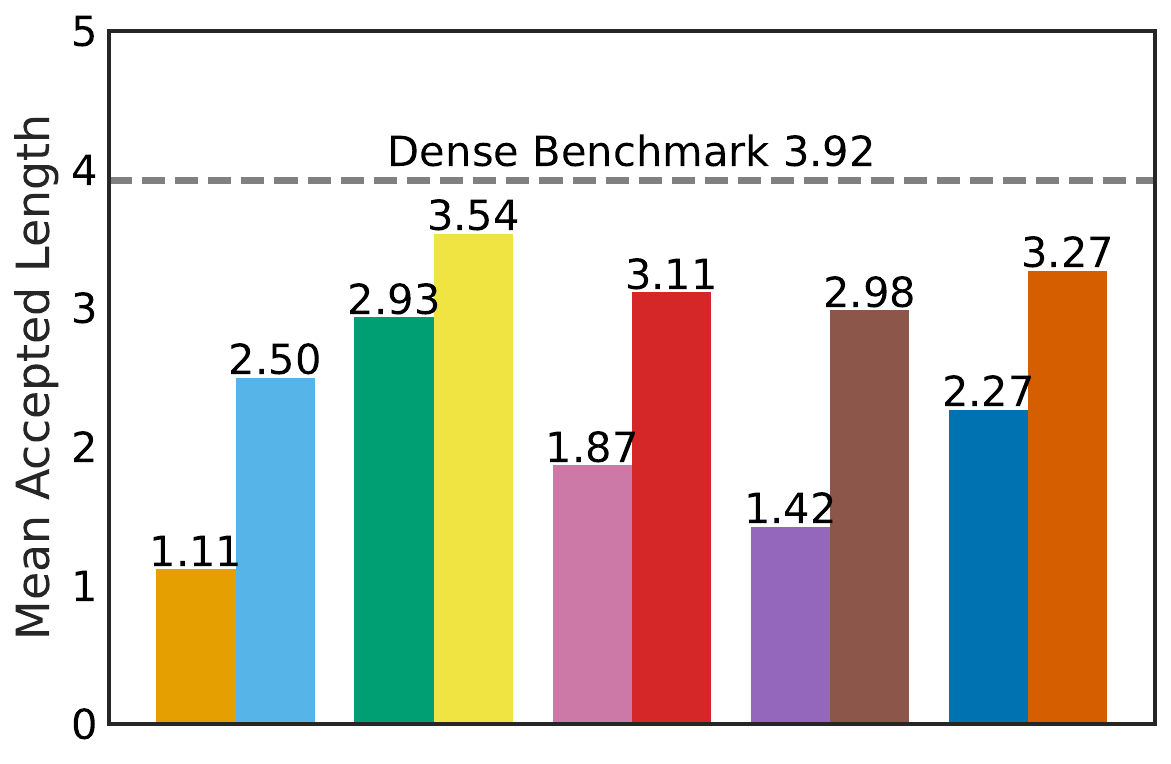}
    \subcaption{Llama-3.2-1B-Instruct}
    \label{fig:llama32-1b-spec}
  \end{minipage}
  \hfill 
  \begin{minipage}[b]{\figwidth}
    \centering
    \includegraphics[width=\figwidth]{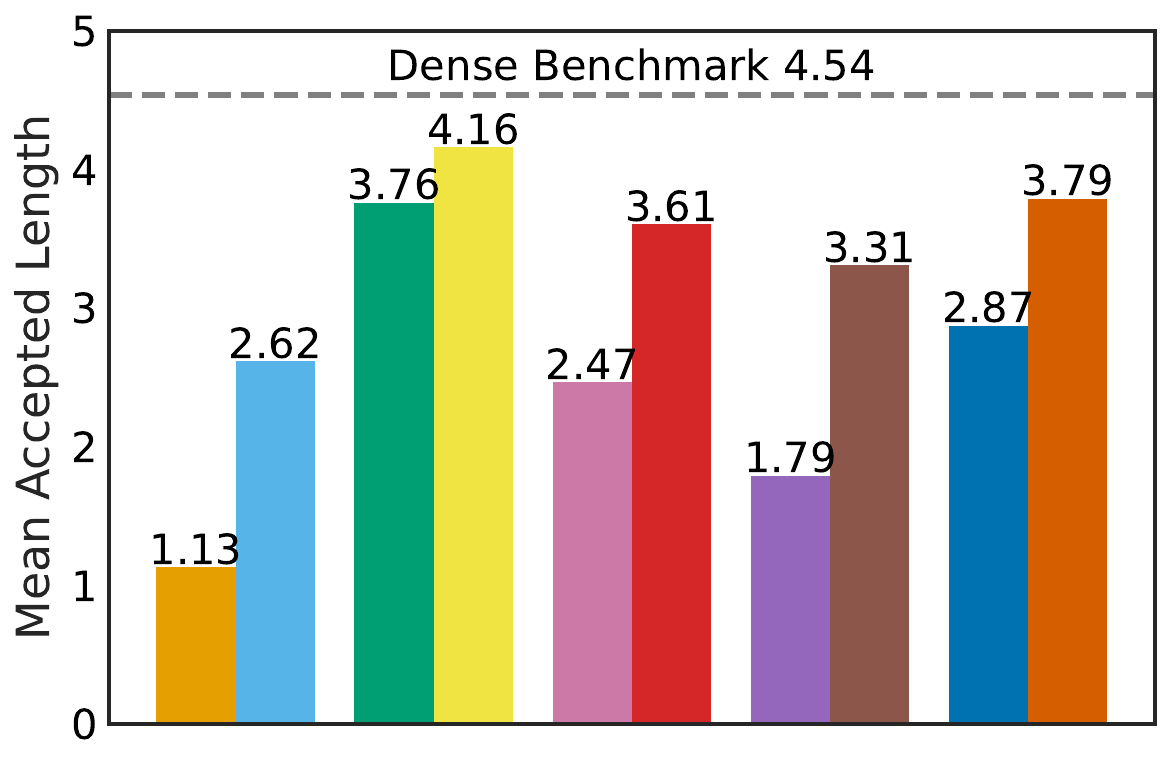} 
    \subcaption{Llama-3.2-3B-Instruct}
    \label{fig:llama32-3b-spec}
  \end{minipage}
  \hfill 
  \begin{minipage}[b]{\figwidth} 
    \centering
    \includegraphics[height=\figheight]{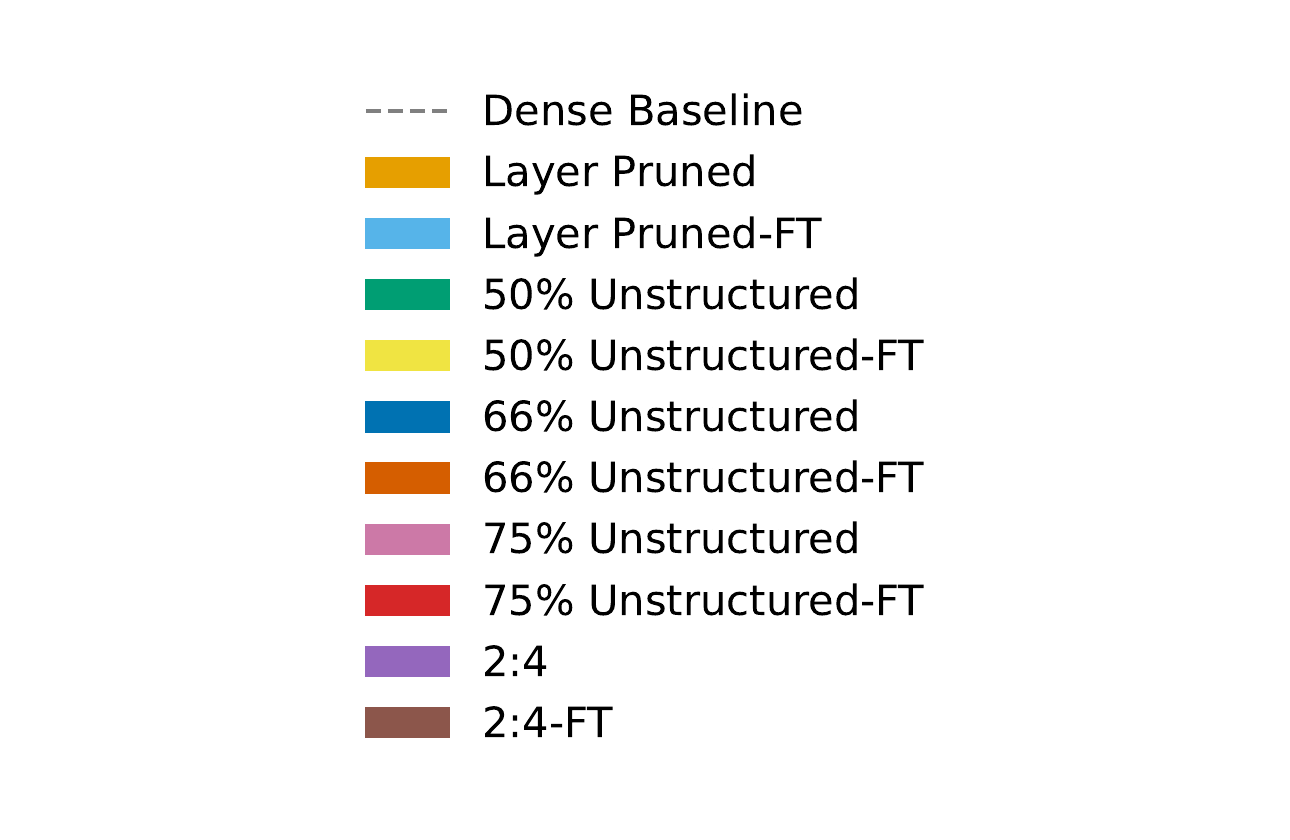} 
    \label{fig:spec_legend}
  \end{minipage}
  \vspace{1em} 
  \begin{minipage}[b]{\figwidth}
    \centering
    \includegraphics[width=\figwidth]{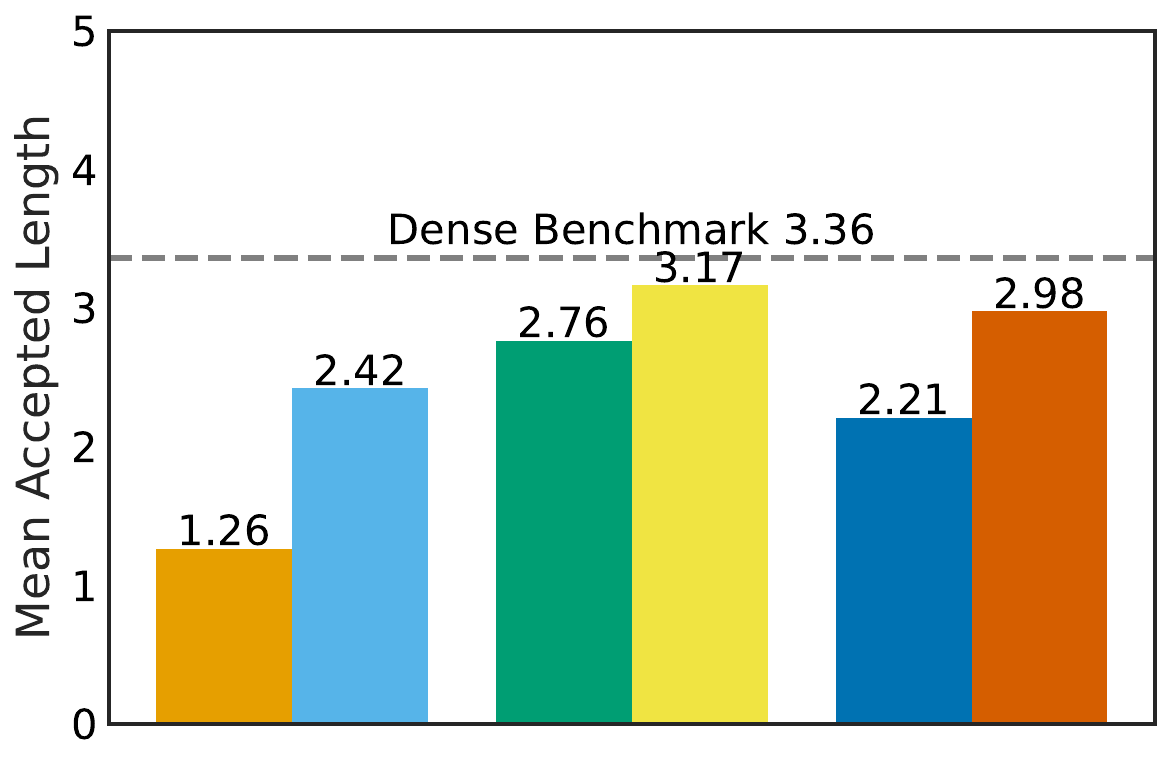}
    \subcaption{Qwen-2.5-0.5B-Instruct}
    \label{fig:qwen25-0.5b-spec}
  \end{minipage}
  \hfill
  \begin{minipage}[b]{\figwidth}
    \centering
    \includegraphics[width=\figwidth]{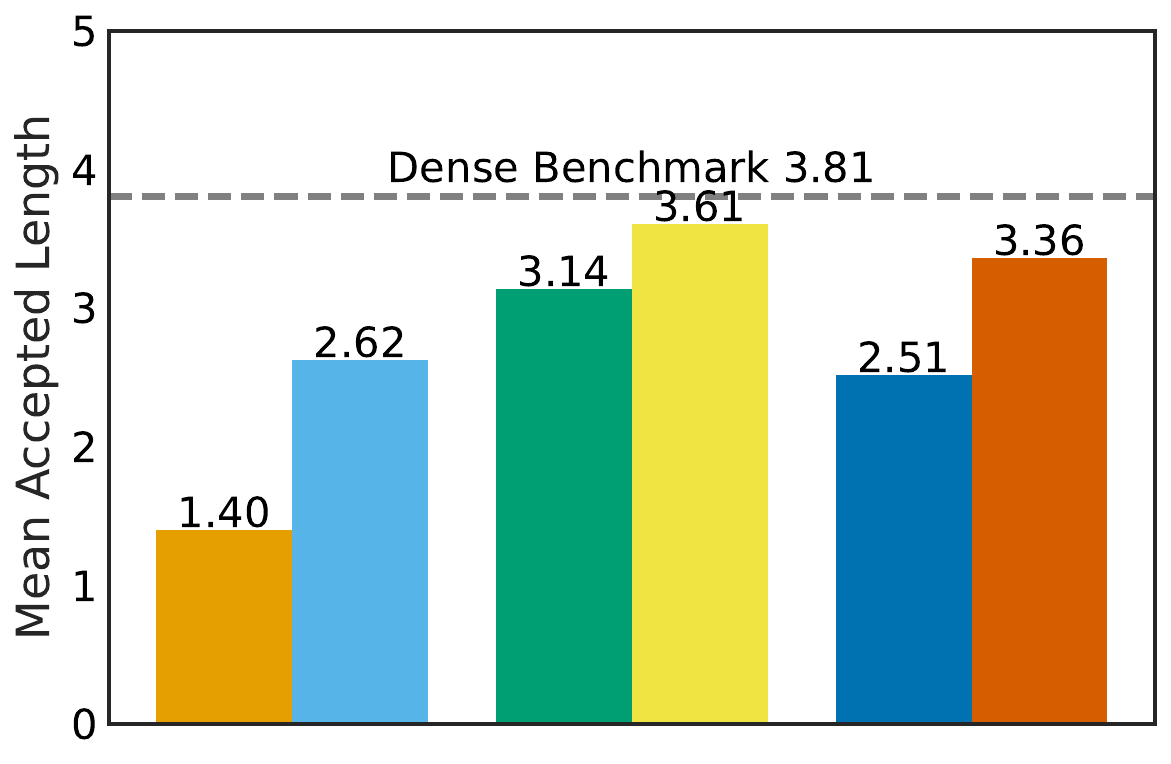} 
    \subcaption{Qwen-2.5-1.5B-Instruct}
    \label{fig:qwen25-1.5b-spec}
  \end{minipage}
  \hfill
  \begin{minipage}[b]{\figwidth}
    \centering
    \includegraphics[width=\figwidth]{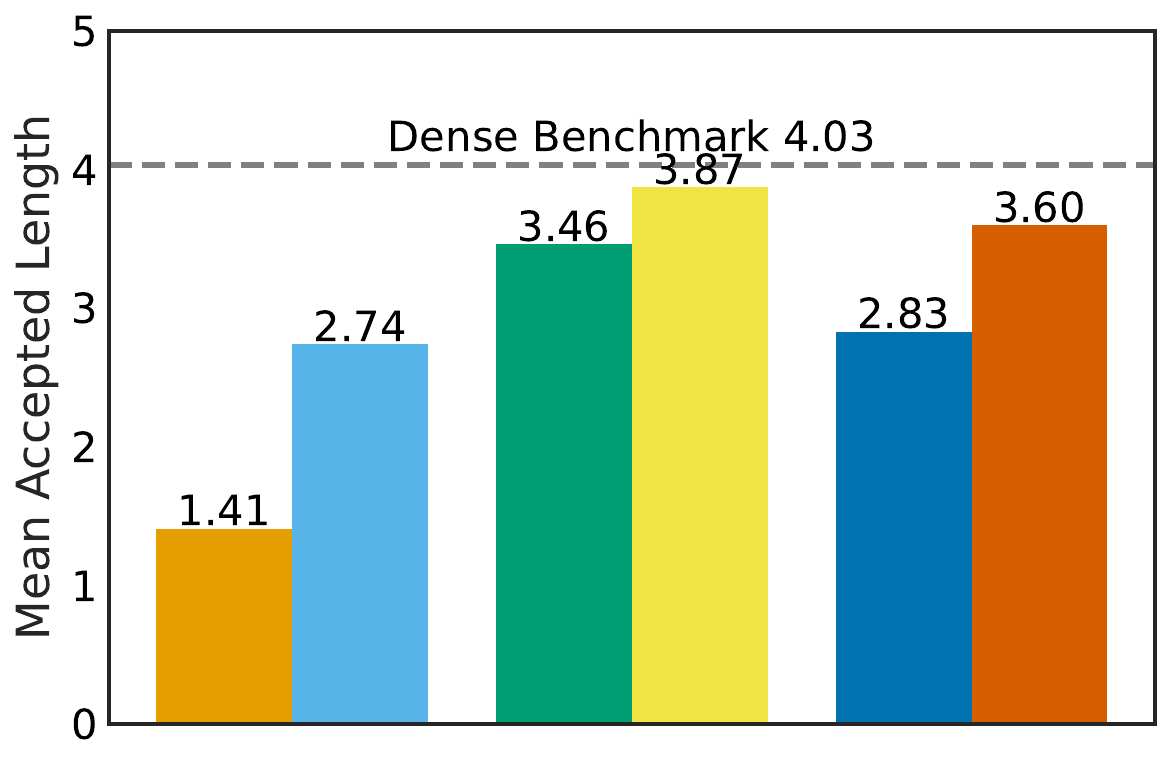} 
    \subcaption{Qwen-2.5-3B-Instruct}
    \label{fig:qwen25-3b-spec}
  \end{minipage}
  
  \caption{\textbf{\gls{mal} for Llama-3.2 and Qwen-2.5 model families} on SpecBench~\citep{xia_unlocking_2024} for layer-pruned, unstructured sparse, and 2:4 sparse draft models including results after one-shot pruning and \gls{sdsd} fine-tuning. Dense model baseline are depicted in horizontal grey lines on each plot. \textbf{Top row:} Llama-3.2 draft models speculating for a Llama-3.1-70B-Instruct target model. \textbf{Bottom row:} Qwen-2.5 draft models speculating for Qwen-2.5-72B-Instruct. Across both model families, we observe that the fine-grained sparse draft models achieve significantly higher \glspl{mal} than the layer-pruned models.}
  \label{fig:spec-bench}
\end{figure*}

\subsubsection{Speculative decoding}
SpecBench~\cite{xia_unlocking_2024} assesses draft token acceptance rate across a variety of tasks including translation, summarization, question answering, mathematical reasoning, multi-turn conversation, and retrieval augmented generation (RAG).

In \cref{fig:spec-bench}, we report the \gls{mal} on SpecBench~\citep{xia_unlocking_2024} for a variety of models and sparsities. We find that the \gls{sdsd} 50\% unstructured drafters achieve the highest \gls{mal} of the sparse models, but do not exceed the dense baselines. For Qwen-2.5, the \gls{sdsd} 50\% unstructured drafter achieve a \gls{mal} within 0.2 of the dense models across all model sizes investigated, despite lower overall scores compared to the Llama model family. We speculate that the overall higher \gls{mal} for the Llama-3.2 draft models are the result of improved alignment with the target model stemming from the pruning and distillation process used during pretraining~\citep{llama3}.

The importance of the draft model alignment is clear; our self-data distillation and fine-tuning process enables much higher \gls{mal} than the one-shot pruned models. Even more strikingly, the \gls{sdsd} effectively aligns the draft model in the \gls{uag} setting \textit{across model families}, as can be seen in \cref{fig:uag-spec}. Remarkably, the sparse Qwen drafters outperform their dense counterparts under this setting, further highlighting the importance of draft model alignment and the benefits of \gls{sdsd}.

\subsection{\glspl{mac} analysis}
The prior section demonstrate that \gls{sdsd} drafters have comparable quality to dense models, it remains unclear whether their efficiency gains yield offer practical benefits. Unstructured sparsity is not well suited for \glspl{gpu} and tangible latency improvements require specialized kernels. An analysis based on \glspl{mac} provides a theoretical limit of the benefits one can expect from unstructured sparsity.

\begin{figure*}[t]
        \centering
        \includegraphics[width=1.0\textwidth]{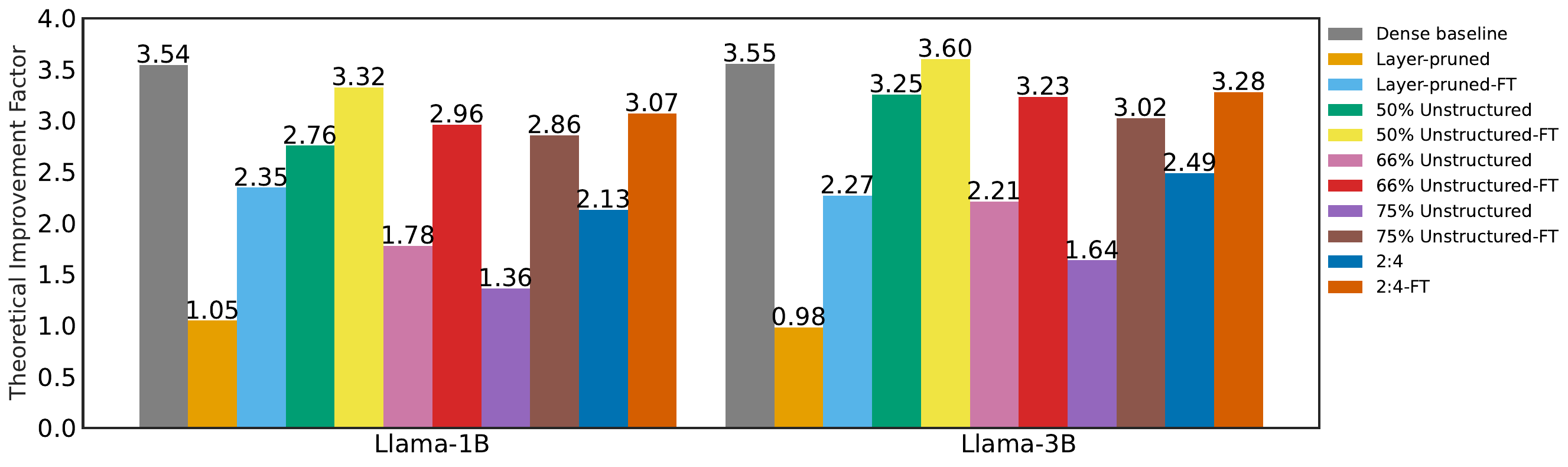}
        \caption{\textbf{Theoretical improvement factor for \glstext{sdsd} Llama-3.2 models} drafting for a Llama-3.1-70B-Instruct based on \glstext{mal} and \glstext{mac}s. Our \glstext{sdsd} Llama-3B draft model offers the highest improvement factor compared to dense or layer-pruned models.} 
        \label{fig:theoretical-improvement-factor}
\end{figure*}

In \cref{fig:macs}, we plot the \gls{mal} of our models trained with \gls{sdsd} versus their corresponding MACs. The unstructured sparse Qwen-2.5 drafters nearly achieve iso-\gls{mac} performance improvements compared to their dense counterparts. However, additional \glspl{mac} associated with the \gls{lm-head} prevent the unstructured sparse models from directly outperforming the dense drafters in terms of \gls{mal} vs. \glspl{mac}. In \cref{fig:theoretical-improvement-factor} we demonstrate that our best performing \gls{sdsd} draft models provide the highest \textit{theoretical} improvement factor of all draft model candidates examined. 

\subsection{Compressed sparse drafters}\label{sec:compressed-drafters}
As noted in \cref{eq:improvement-factor}, a crucial property to consider when evaluating speculative decoding draft models is $c$, the ratio of the draft to target model latency. To realize practical benefits of \gls{sdsd} on commodity hardware available today, we require efficient sparse representations and kernels which offer lower latency than the dense baselines. To this end, we benchmark our \gls{sdsd} unstructured draft models using nm-vLLM~\citep{NeuralMagic2024nmvllm,kwon2023efficient} and compare them with dense and layer-pruned draft models in \cref{fig:nm_vllm_combined}. \gls{sdsd} drafters outperform layer-pruned models and even dense drafters in the 3B model size category. While the smallest dense model in each model family achieves a higher overall improvement factor, these results highlight the potential for sparsity to enable the use of larger draft models. As foundation model sizes continue to grow, increasing the capacity of draft models may be necessary to maintain sufficient \gls{mal} and support novel use-cases such as scaling test-time compute. See \cref{sec:benchmarking-details} for details on our benchmarking methodology.

\section{Related work}\label{sec:related-work}
Speculative decoding with greedy verification was initially proposed by \citet{stern_blockwise_2018}. Speculative sampling was introduced concurrently by \citet{leviathan_fast_2023} and \citet{chen_accelerating_2023}. Since these initial investigations, several works have proposed modifications and refinements to the original framework. In Medusa~\citep{cai_medusa_2024}, the authors reformulated used a single model with multiple lm-head layers, each of which predicts multiple draft token candidates across each draft token position, introducing the first instance of draft \textit{token-trees}. Specifically related to our work, Medusa also used self-distillation to obtain a fine-tuning dataset. \citet{chen_magicdec_2024} examined speculative decoding for long context lengths, finding that constant KV-cache sizes lead to improved acceleration. EAGLE~\citep{li_eagle_2024,li_eagle-2_2024} proposed an efficient draft model consisting of a single layer auto-regression head trained from scratch and the frozen embedding and lm-head layers from the target model. Token trees were further investigated by \citet{miao_specinfer_2024} in which a parallel decoding algorithm was also proposed. \citet{sun_triforce_2024} introduced a hierarchical framework in which efficient KV-cache implementations are used for drafting. Speculative decoding was combined with early-exit mechanisms for drafting by \citet{zhang_draft_2024} and \citet{liu_kangaroo_2024}. We note that optimizations such as draft token-trees and KV-cache compression are orthogonal to \gls{sdsd} and future work could consider their integration with our method.

The surprising result that a significant fraction of decoder blocks can be pruned from \glspl{llm} without incurring excessive quality reduction was examined in detail in several works~\citep{jha_just_2024,sun_transformer_2024,yang_laco_2024,men_shortgpt_2024}. In particular, these works found that the ``middle'' decoder blocks are the most amenable to pruning. However, abstract reasoning tasks were noted to be disproportionality affected by layer-pruning. We specifically highlight \citet{gromov_unreasonable_2024} which introduced the angular cosine distance metric that is used for our layer-pruned baselines.  

Fine-grained sparsity for \glspl{llm} has received significant attention from the research community. Efficient one-shot pruning methods such as SparseGPT~\citep{frantar_sparsegpt_2023}, Wanda~\citep{sun_simple_2023}, and Plug-and-Play~\citep{zhang_plug-and-play_2023} proposed re-framing the pruning procedure as a layer-wise reconstruction of the original model weights, using the magnitudes and/or activations to determine the saliency of individual weights. \citet{fang_maskllm_2024} proposed freezing the model weights and training the mask alone to efficiently obtain 2:4 sparse \glspl{llm}. \gls{owl}~\citep{yin_outlier_2023} and AlphaPruning~\citep{lu_alphapruning_2024} proposed non-uniform layer-wise sparsity distributions which outperform uniform distributions in terms of perplexity and downstream task accuracy. Despite these efforts, sparse \glspl{llm} still fall short of their dense counterparts in many respects~\citep{jaiswal_compressing_2023,thangarasa_spdf_2023,yin_junk_2023}.

\section{Conclusion}\label{sec:conclusion}
We introduce the \gls{sdsd} methodology for obtaining fine-grained sparse draft models. \gls{sdsd} consists of self-data distillation, one-shot pruning, and sparse fine-tuning. We find that self-data distillation effectively aligns draft models, even if their target is from a different model family such as in the \gls{uag} setting. \gls{sdsd} shows significant theoretical benefits based \glspl{mac}. When paired with efficient sparse kernels, \gls{sdsd} outperforms layer-pruned models at all model sizes and compression ratios evaluated. For the 3B draft models, \gls{sdsd} improves end-to-end speculative decoding latency compared to dense draft models, revealing the potential for using larger drafters. See \cref{sec:limitations} for a discussion of limitations and future work. We hope this work will inspire further efforts to develop low-latency sparse kernels and encourage software/hardware co-design for efficient \gls{llm} inference. 

\section{Acknowledgements}
We would like to acknowledge the helpful feedback of Ganesh Venkatesh during our initial explorations that led to this work. ML and YI gratefully acknowledge the support of Alberta Innovates (ALLRP-577350-22, ALLRP-222301502), the Natural Sciences and Engineering Research Council of Canada (NSERC) (RGPIN-2022-03120, DGECR-2022-00358), and Defence Research and Development Canada (DGDND-2022-03120). YI is supported by a Schulich Research Chair.

\newpage

\bibliography{bib}
\bibliographystyle{acl_natbib}

\newpage
\appendix
\onecolumn

\section{Impact Statement}\label{sec:imapact}
Training and serving \glspl{llm} consume large amounts of energy, even for efficient implementations that leverage techniques discussed in this paper. Speculative decoding requires more VRAM than auto-regressive sampling, necessitating the use of additional hardware devices. There are many potential negative societal consequences of generative AI, we hope that by improving the efficiency of these models we will help distribute the benefits of such systems more equitably to the broader public. 

\section{Limitations and future work}\label{sec:limitations}
The effectiveness of \gls{sdsd} in improving inference efficiency depends on software and hardware optimized for sparsity. The nm-vLLM unstructured sparse kernel we use to benchmark is optimized for Ampere-series NVIDIA \glspl{gpu}, and evaluating its performance on alternative hardware, such as TPUs or custom accelerators, is crucial for broader adoption. While \gls{sdsd} requires significant compute for fine-tuning and self-data distillation, this investment could further enhance the quality of dense models, particularly in speculative decoding. Several promising research directions emerge from this work. One is integrating quantization-aware training~\citep{chen2024efficientqatefficientquantizationawaretraining}, draft-token trees, and compressed KV-cache implementations~\citep{shi2024costdownreviewmethods} with \gls{sdsd} to improve memory and compute efficiency. Another is benchmarking \gls{sdsd} on hardware optimized for unstructured sparsity, such as Cerebras' wafer-scale engine. Incorporating advanced fine-tuning strategies like speculative knowledge distillation~\citep{xu_speculative_2024} or square-head distillation~\citep{kurtic_sparse_2023} could further improve sparse draft model quality. Finally, combining structured pruning with fine-grained sparsity and quantization presents an opportunity to reduce inference-time latency while maintaining accuracy, making sparsity-aware methods like \gls{sdsd} more widely applicable across diverse hardware architectures.

\section{Hyperparameter settings and implementation 
details}\label{sec:hyperparameters}
The core components of \gls{sdsd} consist of: one-shot pruning, self-data distillation, and sparse fine-tuning. 

\subsection{One-shot pruning}
For one-shot pruning, we use the default SparseGPT hyperparameters. Explicitly, we use 0.01, 128, and 16 bits for the Hessian damping, block size, and model precision, respectively. For calibration, we randomly select 128 samples from a subset\footnote{\url{https://huggingface.co/datasets/robbiegwaldd/dclm-micro}} of the DCLM dataset~\citep{li2024datacomplmsearchgenerationtraining} with a sequence length of 2048 tokens. We selected the DCLM calibration dataset based on the results of \citet{ji_beware_2024}. 

We prune our draft model candidates to 0.5, 0.66, and 0.75 unstructured sparsity in addition to 2:4 sparsity. For our experiments with non-uniform layer-wise sparsity distributions, we explore both \gls{owl} and our proposed angular distance layer-wise distribution, the results of which are presented in \cref{sec:pruning-results}. 

However, despite apparent advantages in terms of perplexity when measured on WikiText-V2~\citep{merity_pointer_2016}, we find that the non-uniform layer-wise sparsity distributions considered offer little benefit to downstream evaluation tasks or \gls{mal} during speculative decoding. Furthermore, since non-uniform layer-wise sparsity distribution pose a challenge for inference systems that leverage pipelining with continuous batching, we opt to use the more straightforward uniform layer-wise sparsity distribution for all results included in \cref{sec:results}.

\subsection{Self-data distillation}
Our self-data distillation follows the original implementation\footnote{\url{https://github.com/sail-sg/sdft}} of ~\citet{yang_self-distillation_2024}. We generate the distilled labels using 16 bit float precision for the model weights and a maximum generation length of 4096 tokens. We sample from the target model with top-p sampling with $p=1.0$ and a temperature of 0.9. Using Llama-3.1-70B, we use self-data distillation to produce aligned fine-tuning datasets for Opus-100 translation~\citep{opus}\footnote{\url{https://huggingface.co/datasets/Helsinki-NLP/opus-100}} and MathInstruct-V2~\citep{toshniwal2024openmath2} datasets. Using Qwen-2.5-72B-Instruct, we distil MathInstruct-V2. 

We create the distillation inputs by combining the original input, original output, and context with the default chat templates. For MathInstruct-V2 we applied the following template:

\begin{lstlisting}[language=Python]
_register_template(
    name="llama3_orig_math_distill",
    format_user=StringFormatter(
        slots=[
            (
                "<|start_header_id|>user<|end_header_id|>\nQuestion:\n\n{{content}}\nAnswer:\n\n{{resp}}\n\nGreat! Let's think step by step.<|eot_id|>"
                "<|start_header_id|>assistant<|end_header_id|>\n\n"
            )
        ]
    ),
    format_system=StringFormatter(slots=["<|start_header_id|>system<|end_header_id|>\n\n{{content}}<|eot_id|>"]),
    format_observation=StringFormatter(
        slots=[
            (
                "<|start_header_id|>tool<|end_header_id|>\n\n{{content}}<|eot_id|>"
                "<|start_header_id|>assistant<|end_header_id|>\n\n"
            )
        ]
    ),
    format_prefix=EmptyFormatter(slots=[{"bos_token"}]),
    default_system=(
        "You are a math expert tasked with generating high-quality responses. You will be provided with a math question and a reference answer. Your goal is to rewrite the reference answer in a clear and accurate manner, ensuring it thoroughly addresses the question. Maintain mathematical rigor while improving clarity where necessary."
    ),
    stop_words=["<|eot_id|>"],
    replace_eos=True,
    replace_jinja_template=False,
)
\end{lstlisting}

For Opus, we applied the following template:

\begin{lstlisting}[language=Python]
_register_template(
    name="llama3_multilingual_distill",
    format_user=StringFormatter(
        slots=[
            (
                "<|start_header_id|>user<|end_header_id|>\n\n{{content}}\n\n{{resp}}<|eot_id|>"
                "<|start_header_id|>assistant<|end_header_id|>\n\n"
            )
        ]
    ),
    format_system=StringFormatter(
        slots=[{"bos_token"}, "<|start_header_id|>system<|end_header_id|>\n\n{{content}}<|eot_id|>"]
    ),
    default_system=(
        "You are a language translation expert tasked with generating high-quality translations. You will be provided with a sentence or passage in the source language and a reference translation. Your goal is to rewrite the reference translation to ensure it is both accurate and fluent, preserving the original meaning while improving clarity and readability. Ensure cultural nuances and context are respected during the translation process."
    ),
    stop_words=["<|eot_id|>"],
    replace_eos=True,
)
\end{lstlisting}

In addition to these datasets, we leverage publicly available datasets produced using the self-synthesis method Magpie~\citep{xu2024magpiealignmentdatasynthesis} for both Llama-3.1-70B-Instruct\footnote{\url{https://huggingface.co/datasets/Magpie-Align/Magpie-Llama-3.1-Pro-MT-300K-Filtered}} and Qwen-2.5-72B-Instruct\footnote{\url{https://huggingface.co/datasets/Magpie-Align/Magpie-Qwen2.5-Pro-300K-Filtered}} target models. As the Magpie datasets are generated by the target models, further alignment with self-data distillation is redundant. 

\subsection{Sparse fine-tuning}
For sparse fine-tuning, we extend Llama-Factory~\cite{Zheng_LlamaFactory_Unified_Efficient_2024} to incorporate our sparse fine-tuning method. We fine-tune our models using a maximum sequence length of 8,192 tokens and truncate any tokens which exceed this limit. For Llama models, we randomly interleave samples from all three fine-tuning datasets with sampling probabilities 75, 12.5, and 12.5\% from Magpie, distilled Opus translation, and distilled MathInstruct-V2, respectively. For Qwen, we randomly interleave samples from Magpie and distilled MathInstruct-V2 with probabilities 80 and 20\%, respectively. 

We optimize our models using the default AdamW optimizer~\citep{adamw} with 1.0e-08, 0.9, and 0.999 for the $\varepsilon$, $\beta_1$, and $\beta_2$ hyperparameters, respectively. For most of our models, we train them with 16,000 optimizer steps with a batch size of 8 for a total of 128,000 samples. The one exception to the above are the 75\% sparse models, which we find benefit from an extended training duration of 32,000 optimizer steps with a batch size of 8 for a total of 256,000 samples. 

We use a grid search optimize the learning rate for all Llama draft model and sparsity combinations. In general, we find that smaller and more sparse models required higher learning rates. Rather than performing a second grid search, we simply re-use the optimal learning rates found for Llama models when fine-tuning Qwen models, using the Llama-3.2-1B-Instruct learning rates for both 0.5B and 1.5B Qwen models. See \cref{tab:lr} for the various learning rates used based on the model size and sparsity type. For learning rate schedule, we use a linear schedule with a linear warm-up using 5\% of the total steps. We also experimented with a cosine schedule but found no significant difference in the validation loss. 

\begin{table}[tb]
    \centering
    \caption{\textbf{Learning rates} used for sparse fine-tuning of the various draft model candidates.}
    \begin{tabular}{c|c|c}
         \textbf{Model} & \textbf{Sparsity} & \textbf{LR}  \\
         \hline
         \multirow{5}{*}{Llama-3.2-1B-Instruct} & 0.5 & 1.25e-05  \\
          & 2:4 & 1.25e-05 \\
          & 50\% layer-pruned & 1.25e-05 \\
          & 0.66 & 2.0e-05 \\
          & 0.75 & 5.0e-05 \\
         \hline
         \multirow{5}{*}{Llama-3.2-3B-Instruct} & 0.5 & 7.5e-06  \\
          & 2:4 & 7.5e-06 \\
          & 50\% layer-pruned & 7.5e-06 \\
          & 0.66 & 1.25e-05 \\
          & 0.75 & 2.0e-05 \\
         \hline
         \multirow{3}{*}{Qwen-2.5-0.5B-Instruct} & 0.5 & 1.25e-05  \\
          & 2:4 & 1.25e-05 \\
          & 50\% layer-pruned & 1.25e-05 \\
         \hline
         \multirow{3}{*}{Qwen-2.5-1.5B-Instruct} & 0.5 & 1.25e-05  \\
          & 2:4 & 1.25e-05 \\
          & 50\% layer-pruned & 1.25e-05 \\
         \hline
         \multirow{3}{*}{Qwen-2.5-3B-Instruct} & 0.5 & 7.5e-06  \\
          & 2:4 & 7.5e-06 \\
          & 50\% layer-pruned & 7.5e-06 \\
    \end{tabular}
    \label{tab:lr}
\end{table}

\subsection{Computational Resources}\label{sec:computational-resources}
We use an internal cluster to run our experiments on a node with four Nvidia H100-80GB-HBM3 \glspl{gpu}, 192 Intel Xeon Platinum 8468 CPUs, and 2TB of RAM. 
Pruning was completed using a single \gls{gpu} in less than 30 minutes per model. Our fine-tuning datasets were generated in approximately 6-8 hours per dataset. For sparse fine-tuning, we use a single \gls{gpu} with the following approximate runtimes: 10 and 21 hours for 1B and 3B, respectively, for the 50\% and 66\% sparse models. The 75\% sparse model fine-tuning took approximately 20 and 42 hours for 1B and 3B, respectively, due to the increased number of optimization steps. 

\section{Non-uniform layer-wise sparsity distributions}\label{sec:pruning-results} 
This section highlights our analysis of non-uniform layer-wise sparsity distributions such as \gls{owl}~\citep{yin_outlier_2023} and a novel distribution inspired by the angular cosine distance measure from ~\citet{gromov_unreasonable_2024}. Ultimately, we found that the non-uniform layer-wise sparsity distributions examined resulted in improved perplexity on WikiText-V2 and a modest improvement on the OpenLLM Leaderboard V1 suite of benchmarks; however, we did not find a statistically significant improvement to \gls{mal} during speculative decoding. 

\gls{owl} allocates sparsity to layers proportional to their \textit{outlier ratio}. The outlier ratio was inspired by the observation that the activations of \glspl{llm} often contain large outliers~\citep{dettmers_llmint8_2022}. Specifically, for a weight matrix $\mathbf{W}$ of shape ($C_{out}$, $C_{in}$) and input $X$, \gls{owl} defines the outlier score of $\mathbf{W}_{i,j}$ as $A_{i,j} = \left \| X_j \right \| \cdot \left | \mathbf{W}_{i,j} \right |$ across all tokens in a calibration dataset. The sparsity ratio is defined as $S^l \propto 1 - D^l \quad S^l \in \{S - \lambda, S+\lambda\}$ where $S \in [0, 1]$ is the uniform sparsity target, $\lambda$ is a hyperparameter which constraints the maximum and minimum layer sparsities, and $D^l$ is the outlier distribution for a single layer calculated as 

\begin{equation}
    D^l = \frac{\sum_{i=1}^{C_{out}}\sum_{j=1}^{C_{in}}
    \mathbf{1}_{\zeta}(A^l_{i,j})}{C_{in}C_{out}}.
\end{equation}

In the above, $\Bar{A}^l$ is the mean of $A^l$, $M$ is a hyperparameter which defines the magnitude of outliers compared to the average activation, and $\mathbf{1}_\zeta$ is the indicator function defined as

\begin{equation}
    \mathbf{1}_\zeta = \begin{cases}
    1 & A_{i,j} > M\cdot \Bar{A}^l \\
    0 & \text{otherwise}.
    \end{cases}
\end{equation}

For our experimental results, we use the default hyperparameters setting $M=5$ and $\lambda = 0.08$.

Our \textit{angular distance layer-wise distribution} is based on the intuition that parameter allocation to decoder blocks should be based on the propensity of each block to modify the residual stream. In essence, decoder blocks which greatly modify their inputs should be allocated more parameters compared to blocks which only slightly modify their inputs. 

Formally, for a decoder with $n$ layers, target global decoder sparsity $S \in [0,1]$, and angular distances $\mathbf{D} \in [D^1, ..., D^N]$ where $D^i$ is defined by \cref{eq:angular-dist} with $n=1$, we allocate the sparsity of each block as

\begin{equation}
    S^i = 1 - \left ( (1-S) * \frac{D^i}{\sum_{j=1}^{n}\left | D^j\right | }
    \right ).
\end{equation}

\setlength{\figwidth}{0.33\textwidth} 
\setlength{\figheight}{0.22\textwidth}
\begin{figure*}[bth]
  \centering
  \begin{minipage}[b]{\figwidth}
    \centering
    \includegraphics[width=\figwidth]{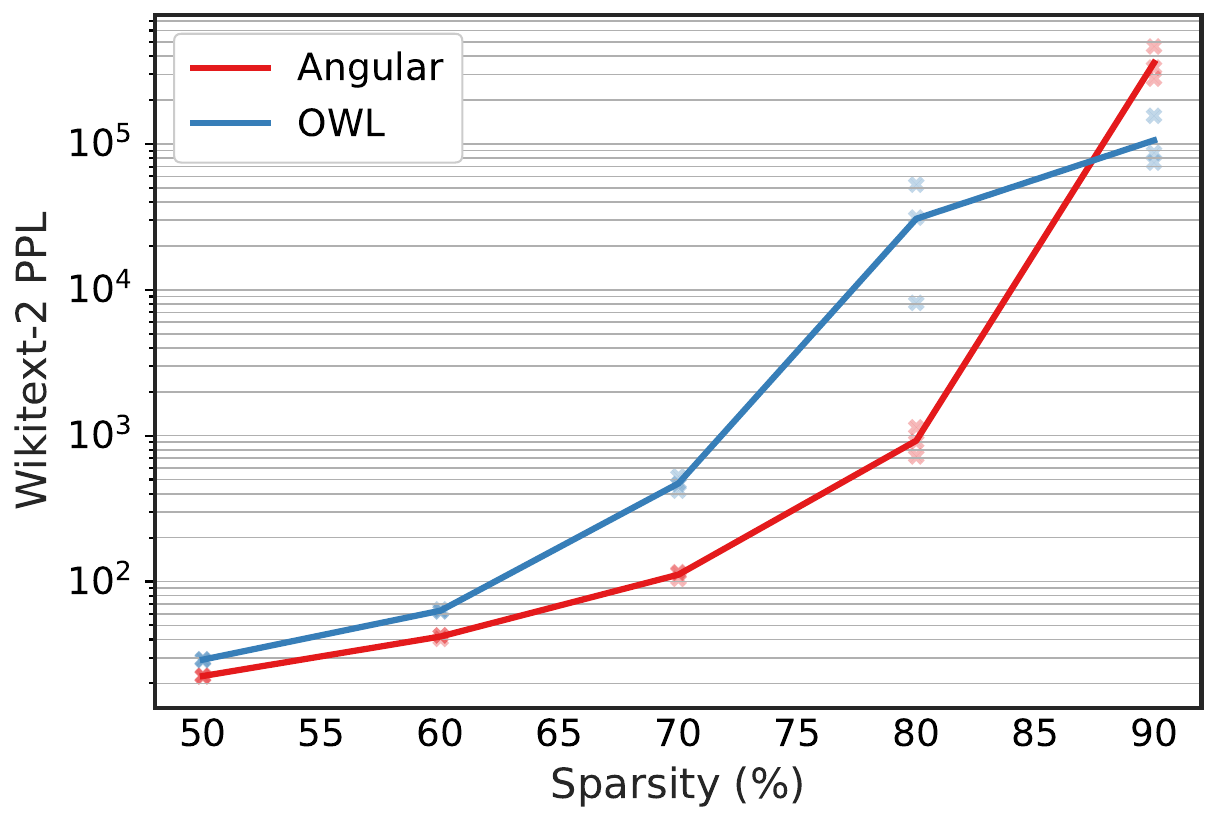} 
    \subcaption{Llama-3.2-1B-Instruct}
    \label{fig:1b-non-uniform-ppl}
  \end{minipage}
  \hfill
  \begin{minipage}[b]{\figwidth}
    \centering
    \includegraphics[width=\figwidth]{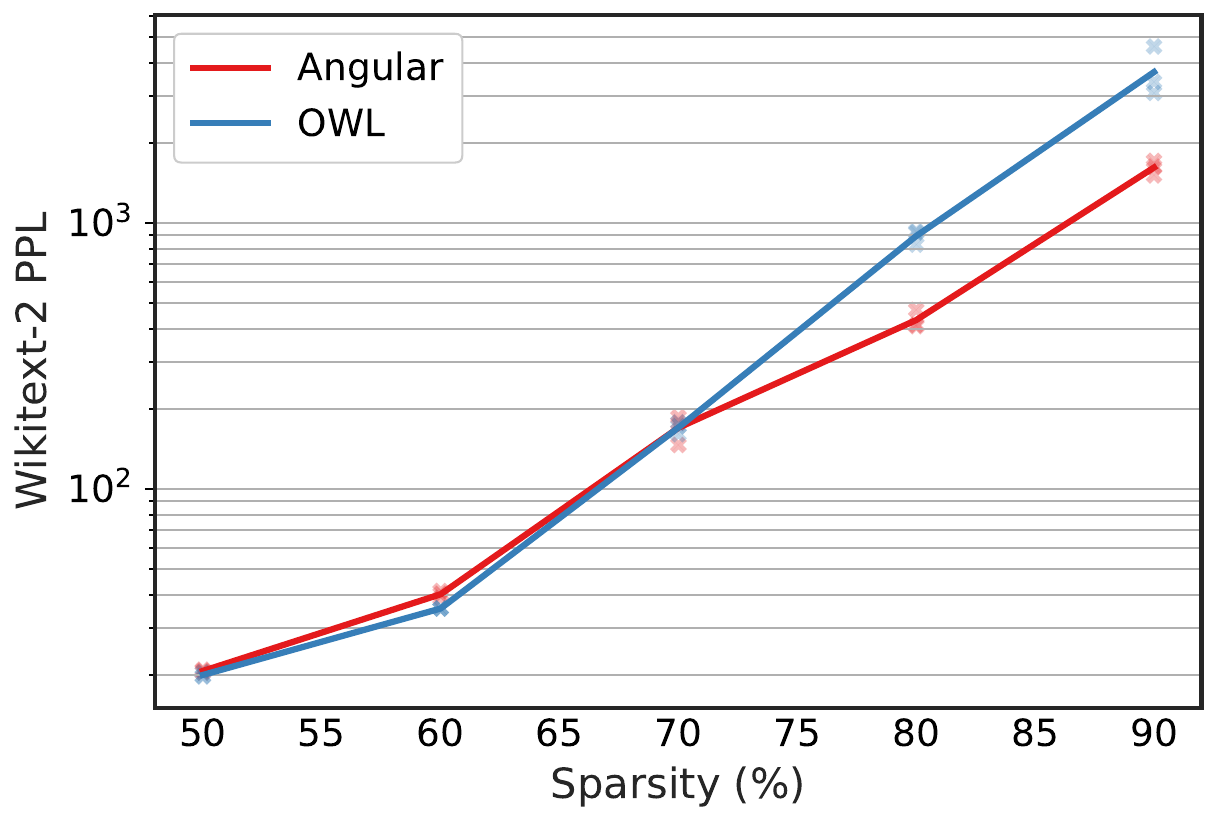} 
    \subcaption{Llama-3.2-3B-Instruct}
    \label{fig:3b-non-uniform-ppl}
  \end{minipage}
  \hfill
  \begin{minipage}[b]{\figwidth}
    \centering
    \includegraphics[width=\figwidth]{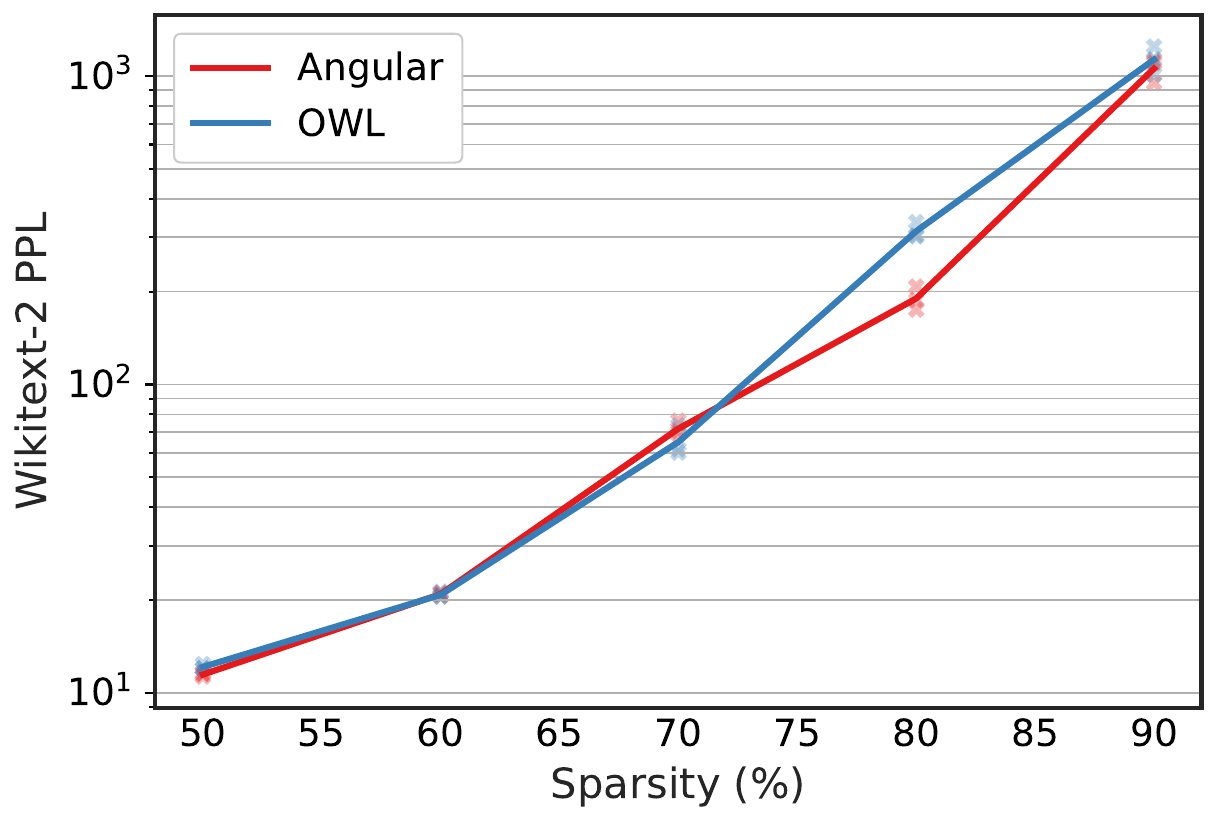} 
    \subcaption{Llama-3.2-8B-Instruct}
    \label{fig:8b-non-uniform-ppl}
  \end{minipage}

  \vspace{1em} 


  \begin{minipage}[b]{\figwidth}
    \centering
    \includegraphics[width=\figwidth]{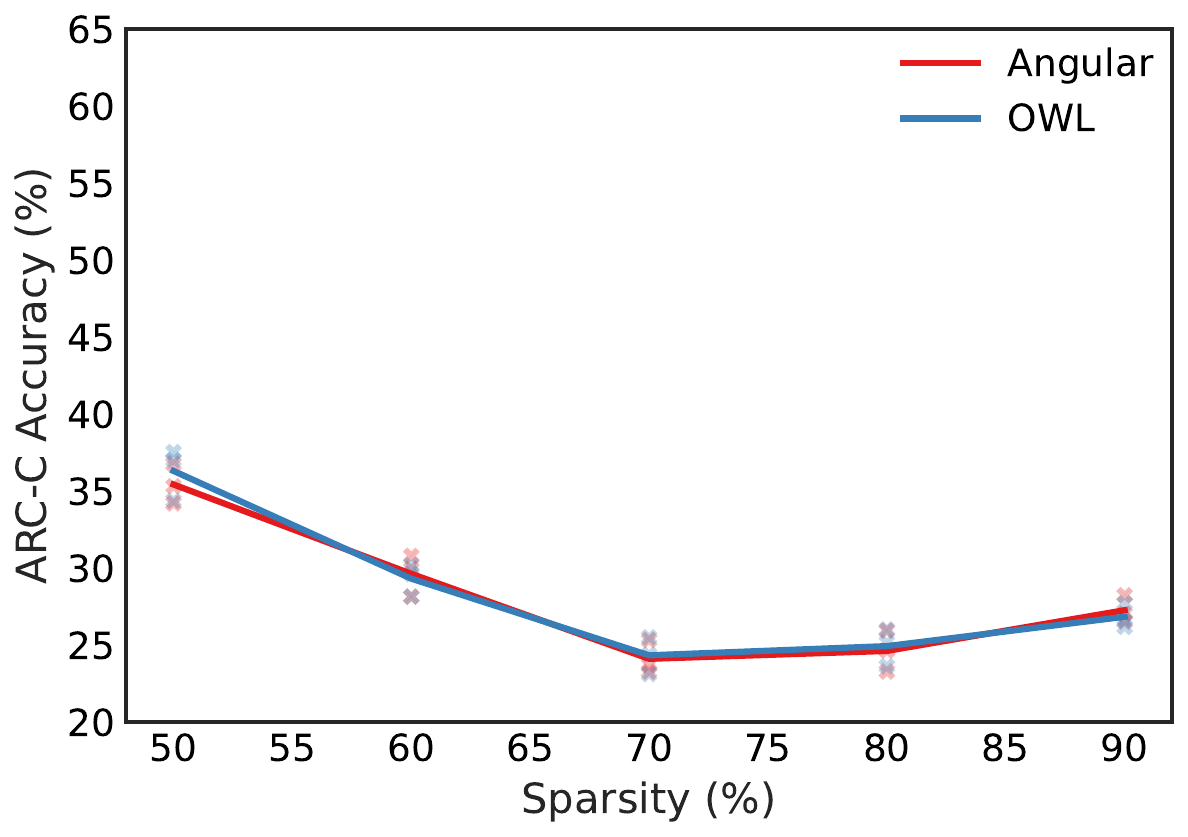}
    \subcaption{Llama-3.2-1B-Instruct}
    \label{fig:1b-non-uniform-arc}
  \end{minipage}
  \hfill
  \begin{minipage}[b]{\figwidth}
    \centering
    \includegraphics[width=\figwidth]{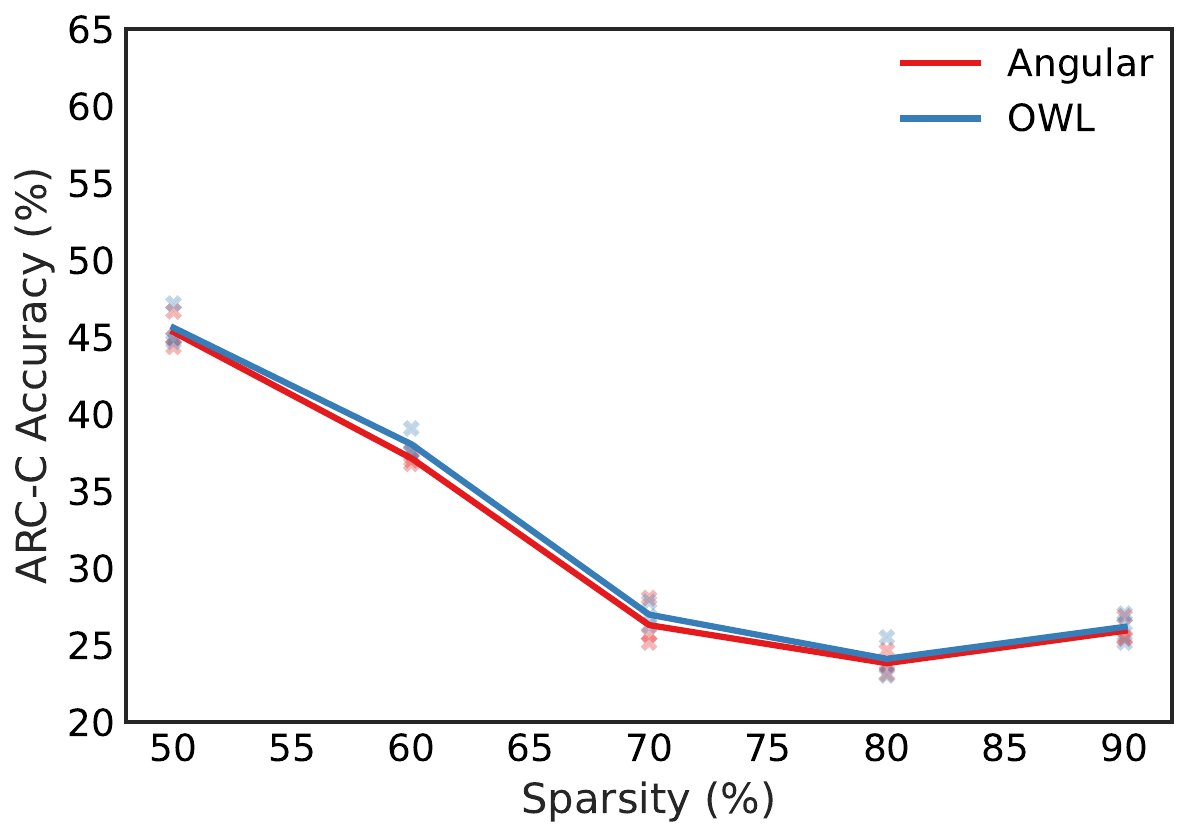} 
    \subcaption{Llama-3.2-3B-Instruct}
    \label{fig:3b-non-uniform-arc}
  \end{minipage}
  \hfill
  \begin{minipage}[b]{\figwidth}
    \centering
    \includegraphics[width=\figwidth]{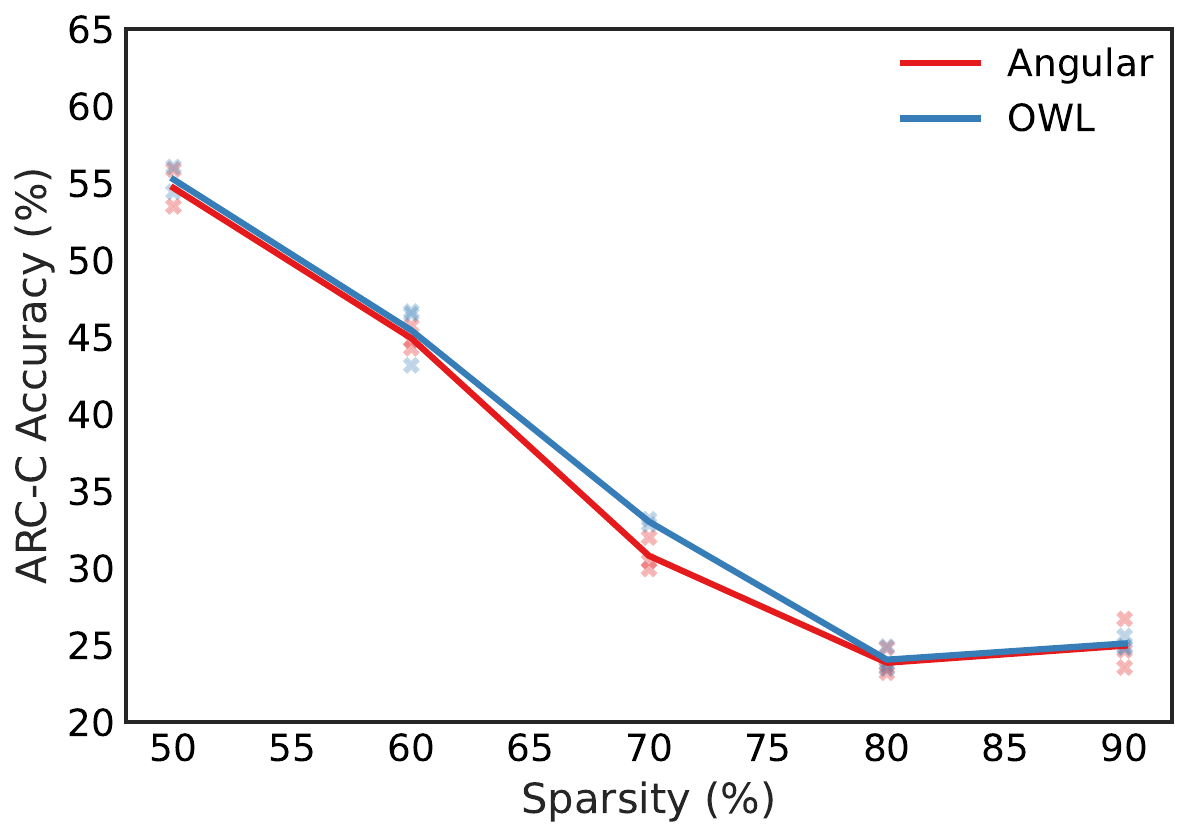} 
    \subcaption{Llama-3.1-8B-Instruct}
    \label{fig:8b-non-uniform-arc}
  \end{minipage}
  \caption{\textbf{Comparison between non-uniform layer-wise sparsity distributions OWL and our proposed angular distance distribution}. In \cref{fig:1b-non-uniform-ppl,fig:3b-non-uniform-ppl,fig:8b-non-uniform-ppl}, we report the WikiText-V2 Perplexity for Llama-3.2 1B and 3B and Llama-3.1 8B at 50, 60, 70, 80, and 90\% sparsity. At high sparsities, the angular distance distribution outperforms OWL, particularly for the 1B model. In \cref{fig:1b-non-uniform-arc,fig:3b-non-uniform-arc,fig:8b-non-uniform-arc}, we report the ARC-C accuracy for the same models and sparsities. Despite outperforming OWL in terms of perplexity, we find that OWL achieves slightly higher scores on this benchmark. Further, neither distribution yields statistically significant improvements to acceptance rate in our limited experiments.}
  \label{fig:non-uniform-dist}
\end{figure*}

\section{Benchmarking details}\label{sec:benchmarking-details}
\begin{figure*}[t]
        \centering
        \includegraphics[width=1.0\textwidth]{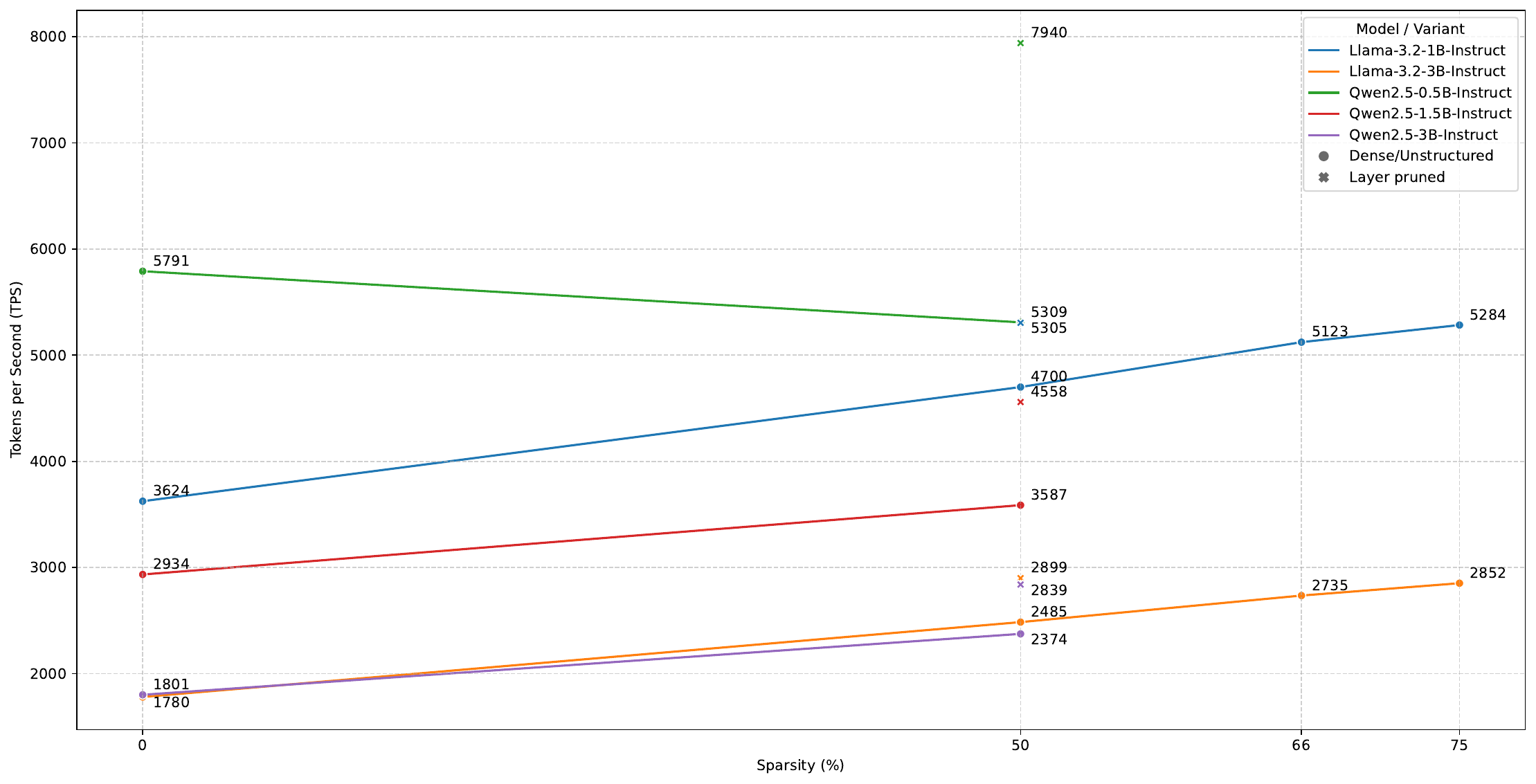}
        \caption{\textbf{Throughput (tok/s) vs. sparsity by model and variant.} Layer-pruned draft models offer ultra-low latency; however, their reduced \gls{mal} results in an overall lower improvement factor compared to \gls{sdsd} unstructured sparse models. All models are benchmarked using nm-vLLM. Throughput values include both input and output tokens.} 
        \label{fig:nm-vllm-throughput}
\end{figure*}

Benchmarking was completed on a 4 $\times$ NVIDIA A6000 \gls{gpu} node with 88 Intel Xeon E5-2699 v4 CPUs using nm-vLLM~\citep{NeuralMagic2024nmvllm} version 0.5.0. The throughput of our draft and target models is measured using a single request of 300 input tokens and 256 output tokens, the average token counts from the SpecBench benchmark. We repeat the request 10 times to minimize variance in the measurement. The cost factor, $c$, is calculated as the ratio of the draft and target model tokens per second including both input and output tokens. Using \gls{mal} from our SpecBench results, we derive the improvement factor per \cref{eq:improvement-factor}. We report the improvement factor and throughput for dense, layer-pruned, and \gls{sdsd} unstructured drafters in \cref{fig:nm_vllm_combined,fig:nm-vllm-throughput}, respectively. The \gls{sdsd} 1.5B and 3B drafters outperform both dense and layer-pruned models in terms of end-to-end latency improvements. 

\begin{figure*}[t]
        \centering
        \includegraphics[width=1.0\textwidth]{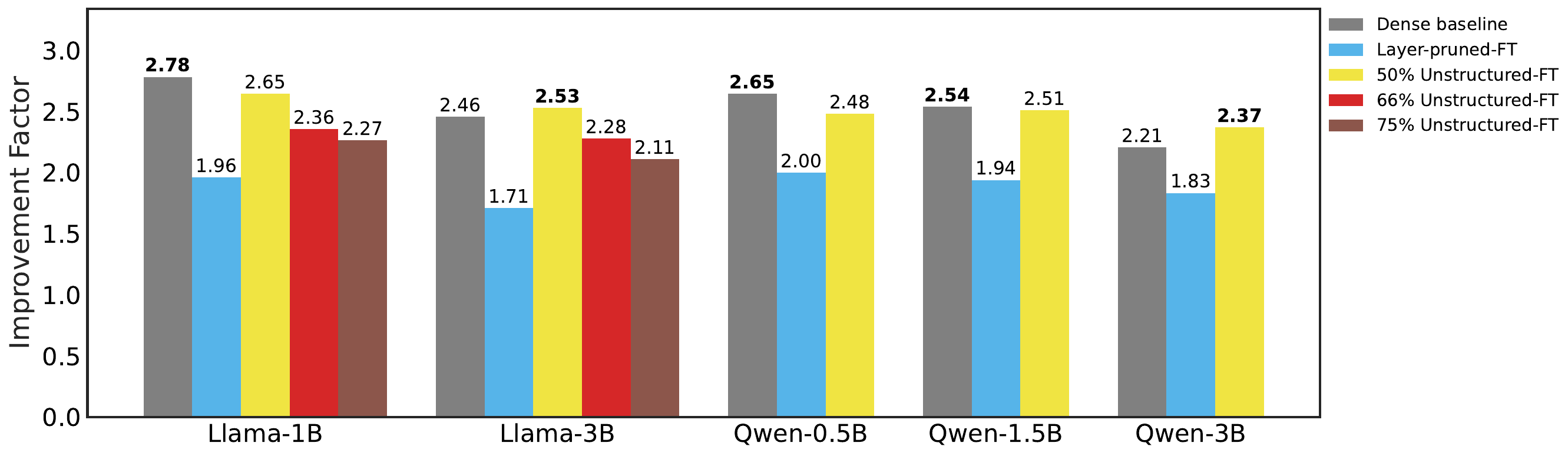}
         \caption{\textbf{Improvement factor of dense, layer-pruned, and \glstext{sdsd} unstructured draft models when benchmarked with 10 concurrent requests}. Llama and Qwen models drafting for Llama-3.1-70B-Instruct and Qwen-2.5-72B-Instruct, respectively. Even in this setting, \glstext{sdsd} drafters outperform layer-pruned draft models and dense drafters in the 3B model size category.} 
        \label{fig:nm_vllm_combined_10_requests}
\end{figure*}

In \cref{fig:nm_vllm_combined_10_requests} we report the improvement factor for 10 \textit{concurrent} requests following a similar benchmarking approach as per the above. Notably, \gls{sdsd} still provides the highest improvement factor for the 3B draft model category, but the 1.5B dense drafter slightly outperforms \gls{sdsd} in this setting. 

\section{Detailed SpecBench results}\label{sec:sb-details}

\begin{figure*}[t]
        \centering
        \includegraphics[width=1.0\textwidth]{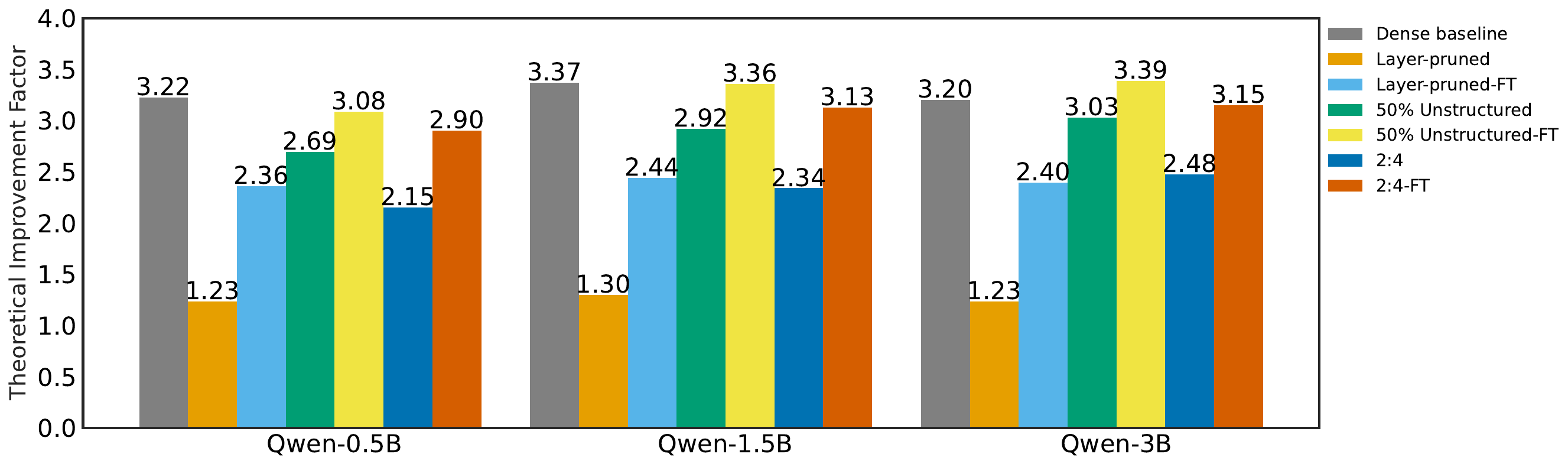}
        \caption{\textbf{Theoretical improvement factor for \gls{sdsd} Qwen-2.5 models} drafting for Qwen-2.5-72B-Instruct based on \gls{mal} and \glspl{mac}. Similar to our Llama results, \gls{sdsd} Qwen-3B offers the highest improvement factor compared to dense or layer-pruned models.} 
        \label{fig:qwen-theoretical-improvement-factor}
\end{figure*}

In this section, we report the detailed SpecBench~\citep{xia_unlocking_2024} results for draft models fine-tuned with \gls{sdsd} (\cref{tab:specbench-details-ft}), one-shot pruned models (\cref{tab:specbench-details-one-shot}) and \gls{sdsd} Qwen-2.5 draft models in the \gls{uag} setting (\cref{tab:specbench-details-uag}). \cref{fig:qwen-theoretical-improvement-factor} depicts the theoretical improvement factor for Qwen-2.5 draft models, in which \gls{sdsd} Qwen-2.5 is found to have the highest improvement factor across all candidate draft models.

\begin{table}[bt]
\centering
\caption{\textbf{SpecBench results for \gls{sdsd} layer-pruned and fine-grained sparse draft models.}}\label{tab:specbench-details-ft}
\begin{tabular}{llcccccccc}
\toprule
 Model & Variant  & \makecell{Sparsity\\\%}  & Overall & \makecell{MT\\Bench} & Translation & Summarization & QA & \makecell{Math\\Reasoning} & RAG \\
\midrule
\multirow[c]{6}{*}{Llama-1B} & Dense & 0 & $3.92$ & $3.99$ & $2.97$ & $3.43$ & $3.55$ & $5.65$ & $3.66$ \\
\cline{2-10}
 & Layer-pruned & 50 & $2.50$ & $2.58$ & $1.61$ & $2.11$ & $2.18$ & $3.91$ & $2.49$ \\
\cline{2-10}
 & \multirow[c]{4}{*}{SparseGPT} & 50 & $3.54$ & $3.68$ & $2.70$ & $3.05$ & $3.00$ & $5.22$ & $3.46$ \\
 &  & 2:4 & $3.27$ & $3.39$ & $2.27$ & $2.87$ & $2.71$ & $5.00$ & $3.21$ \\
 &  & 66 & $3.11$ & $3.23$ & $2.16$ & $2.73$ & $2.57$ & $4.92$ & $3.01$ \\
 &  & 75 & $2.98$ & $3.12$ & $1.97$ & $2.50$ & $2.49$ & $4.84$ & $3.01$ \\
\cline{1-10} \cline{2-10}
\multirow[c]{6}{*}{Llama-3B} & Dense & 0 & $4.54$ & $4.58$ & $3.59$ & $4.12$ & $4.26$ & $5.88$ & $4.27$ \\
\cline{2-10}
 & Layer-pruned & 50 & $2.62$ & $2.70$ & $1.80$ & $2.26$ & $2.25$ & $4.02$ & $2.65$ \\
\cline{2-10}
 & \multirow[c]{4}{*}{SparseGPT} & 50 & $4.16$ & $4.29$ & $3.30$ & $3.74$ & $3.62$ & $5.60$ & $3.94$ \\
 &  & 2:4 & $3.79$ & $3.91$ & $2.98$ & $3.46$ & $3.15$ & $5.42$ & $3.61$ \\
 &  & 66 & $3.61$ & $3.74$ & $2.70$ & $3.21$ & $2.97$ & $5.30$ & $3.55$ \\
 &  & 75 & $3.31$ & $3.47$ & $2.27$ & $2.85$ & $2.71$ & $5.14$ & $3.24$ \\
\cline{1-10} \cline{2-10}
\multirow[c]{5}{*}{Qwen-0.5B} & Dense & 0 & $3.36$ & $3.44$ & $2.99$ & $2.81$ & $2.58$ & $5.70$ & $2.85$ \\
\cline{2-10}
 & Layer-pruned & 50 & $2.42$ & $2.51$ & $1.44$ & $1.91$ & $2.00$ & $3.96$ & $2.09$ \\
\cline{2-10}
 & \multirow[c]{2}{*}{SparseGPT} & 50 & $3.17$ & $3.28$ & $2.15$ & $2.65$ & $2.45$ & $5.15$ & $2.71$ \\
 &  & 2:4 & $2.98$ & $3.09$ & $1.80$ & $2.50$ & $2.32$ & $4.87$ & $2.55$ \\
\cline{1-10} \cline{2-10}
\multirow[c]{5}{*}{Qwen-1.5B} & Dense & 0 & $3.81$ & $3.90$ & $3.72$ & $3.21$ & $2.96$ & $6.02$ & $3.26$ \\
\cline{2-10}
 & Layer-pruned & 50 & $2.62$ & $2.73$ & $1.59$ & $2.08$ & $2.16$ & $4.16$ & $2.31$ \\
\cline{2-10}
 & \multirow[c]{2}{*}{SparseGPT} & 50 & $3.61$ & $3.75$ & $2.83$ & $3.07$ & $2.82$ & $5.35$ & $3.13$ \\
 &  & 2:4 & $3.36$ & $3.50$ & $2.35$ & $2.84$ & $2.58$ & $5.21$ & $2.90$ \\
\cline{1-10} \cline{2-10}
\multirow[c]{6}{*}{Qwen-3B} & Dense & 0 & $4.03$ & $4.17$ & $3.86$ & $3.45$ & $3.15$ & $6.09$ & $3.43$ \\
\cline{2-10}
 & Layer-pruned & 50 & $2.74$ & $2.87$ & $1.58$ & $2.16$ & $2.24$ & $4.38$ & $2.37$ \\
\cline{2-10}
 & \multirow[c]{2}{*}{SparseGPT} & 50 & $3.87$ & $4.04$ & $3.47$ & $3.36$ & $2.97$ & $5.46$ & $3.41$ \\
 &  & 2:4 & $3.60$ & $3.73$ & $3.03$ & $3.13$ & $2.76$ & $5.36$ & $3.13$ \\
\cline{1-10} \cline{2-10}
\bottomrule
\end{tabular}
\end{table}

\begin{table}[bt]
\centering
\caption{\textbf{SpecBench results for one-shot pruned layer-pruned and fine-grained sparse draft models.}}\label{tab:specbench-details-one-shot}
\begin{tabular}{llcccccccc}
\toprule
 Model & Variant  & \makecell{Sparsity\\\%}  & Overall & \makecell{MT\\Bench} & Translation & Summarization & QA & \makecell{Math\\Reasoning} & RAG \\
\midrule
\multirow[c]{6}{*}{Llama-1B} & Dense & 0 & $3.92$ & $3.99$ & $2.97$ & $3.43$ & $3.55$ & $5.65$ & $3.66$ \\
\cline{2-10}
 & Layer-pruned & 50 & $1.11$ & $1.12$ & $1.05$ & $1.09$ & $1.11$ & $1.15$ & $1.11$ \\
\cline{2-10}
 & \multirow[c]{4}{*}{SparseGPT} & 50 & $2.93$ & $3.00$ & $2.12$ & $2.68$ & $2.42$ & $4.52$ & $2.82$ \\
 &  & 2:4 & $2.27$ & $2.26$ & $1.58$ & $2.23$ & $1.91$ & $3.29$ & $2.33$ \\
 &  & 66 & $1.87$ & $1.86$ & $1.29$ & $1.84$ & $1.65$ & $2.47$ & $1.85$ \\
 &  & 75 & $1.42$ & $1.41$ & $1.14$ & $1.40$ & $1.37$ & $1.64$ & $1.43$ \\
\cline{1-10} \cline{2-10}
\multirow[c]{6}{*}{Llama--3B} & Dense & 0 & $4.54$ & $4.58$ & $3.59$ & $4.12$ & $4.26$ & $5.88$ & $4.27$ \\
\cline{2-10}
 & Layer-pruned & 50 & $1.13$ & $1.13$ & $1.07$ & $1.10$ & $1.14$ & $1.20$ & $1.12$ \\
\cline{2-10}
 & \multirow[c]{4}{*}{SparseGPT} & 50 & $3.76$ & $3.82$ & $2.79$ & $3.54$ & $3.17$ & $5.35$ & $3.61$ \\
 &  & 2:4 & $2.87$ & $2.85$ & $2.04$ & $2.92$ & $2.32$ & $4.35$ & $2.93$ \\
 &  & 66 & $2.47$ & $2.44$ & $1.72$ & $2.51$ & $2.02$ & $3.65$ & $2.54$ \\
 &  & 75 & $1.79$ & $1.75$ & $1.27$ & $1.86$ & $1.62$ & $2.23$ & $1.88$ \\
\cline{1-10} \cline{2-10}
\cline{2-10}
 & Layer-pruned & 50 & $1.26$ & $1.26$ & $1.17$ & $1.20$ & $1.24$ & $1.38$ & $1.25$ \\
\cline{2-10}
 & \multirow[c]{2}{*}{SparseGPT} & 50 & $2.76$ & $2.79$ & $2.19$ & $2.51$ & $2.17$ & $4.32$ & $2.41$ \\
 &  & 2:4 & $2.21$ & $2.21$ & $1.63$ & $2.18$ & $1.84$ & $2.90$ & $2.04$ \\
\cline{1-10} \cline{2-10}
\multirow[c]{4}{*}{Qwen-1.5B} & Dense & 0 & $3.81$ & $3.90$ & $3.72$ & $3.21$ & $2.96$ & $6.02$ & $3.26$ \\
\cline{2-10}
 & Layer-pruned & 50 & $1.40$ & $1.39$ & $1.27$ & $1.33$ & $1.35$ & $1.52$ & $1.39$ \\
\cline{2-10}
 & \multirow[c]{2}{*}{SparseGPT} & 50 & $3.14$ & $3.18$ & $2.84$ & $2.87$ & $2.47$ & $4.69$ & $2.75$ \\
 &  & 2:4 & $2.51$ & $2.53$ & $1.96$ & $2.43$ & $2.08$ & $3.41$ & $2.24$ \\
\cline{1-10} \cline{2-10}
\multirow[c]{4}{*}{Qwen-3B} & Dense & 0 & $4.03$ & $4.17$ & $3.86$ & $3.45$ & $3.15$ & $6.09$ & $3.43$ \\
\cline{2-10}
 & Layer-pruned & 50 & $1.41$ & $1.41$ & $1.26$ & $1.31$ & $1.35$ & $1.56$ & $1.38$ \\
\cline{2-10}
 & \multirow[c]{2}{*}{SparseGPT} & 50 & $3.46$ & $3.51$ & $3.27$ & $3.13$ & $2.74$ & $5.12$ & $3.04$ \\
 &  & 2:4 & $2.83$ & $2.82$ & $2.63$ & $2.78$ & $2.30$ & $3.88$ & $2.54$ \\
\cline{1-10} \cline{2-10}
\bottomrule
\end{tabular}
\end{table}

\begin{table}[bt]
\centering
\caption{\textbf{SpecBench results for \gls{sdsd} layer-pruned and fine-grained sparse draft Qwen models drafting for Llama-3.1-70B-Instruct in the \gls{uag} setting.}}\label{tab:specbench-details-uag}
\begin{tabular}{llcccccccc}
\toprule
 Model & Variant  & \makecell{Sparsity\\\%}  & Overall & \makecell{MT\\Bench} & Translation & Summarization & QA & \makecell{Math\\Reasoning} & RAG \\
\midrule
\multirow[c]{3}{*}{Qwen-0.5B} & Dense & 0 & $2.97$ & $3.09$ & $2.04$ & $2.71$ & $2.48$ & $4.24$ & $2.70$ \\
\cline{2-10}
 & \multirow[c]{2}{*}{SparseGPT} & 50 & $3.03$ & $3.18$ & $2.04$ & $2.68$ & $2.52$ & $4.44$ & $2.78$ \\
 &  & 2:4 & $2.87$ & $3.01$ & $1.77$ & $2.54$ & $2.38$ & $4.35$ & $2.68$ \\
\cline{1-10} \cline{2-10}
\multirow[c]{3}{*}{Qwen-1.5B} & Dense & 0 & $3.34$ & $3.47$ & $2.34$ & $3.09$ & $2.82$ & $4.59$ & $2.97$ \\
\cline{2-10}
 & \multirow[c]{2}{*}{SparseGPT} & 50 & $3.42$ & $3.58$ & $2.28$ & $3.08$ & $2.92$ & $4.70$ & $3.04$ \\
 &  & 2:4 & $3.22$ & $3.40$ & $2.04$ & $2.89$ & $2.69$ & $4.59$ & $2.88$ \\
\cline{1-10} \cline{2-10}
\multirow[c]{3}{*}{Qwen-3B} & Dense & 0 & $3.43$ & $3.58$ & $2.40$ & $3.25$ & $2.91$ & $4.55$ & $2.92$ \\
\cline{2-10}
 & \multirow[c]{2}{*}{SparseGPT} & 50 & $3.63$ & $3.82$ & $2.39$ & $3.36$ & $3.05$ & $4.83$ & $3.28$ \\
 &  & 2:4 & $3.44$ & $3.60$ & $2.28$ & $3.17$ & $2.87$ & $4.73$ & $3.13$ \\
\cline{1-10} \cline{2-10}
\bottomrule
\end{tabular}
\end{table}


\section{Detailed OpenLLM Leaderboard V1 results}\label{sec:eval-details}
In \cref{tab:fine-tuned-eval,tab:one-shot-eval}, we report the benchmark results for our model's after fine-tuning and after one-shot pruning, respectively. In \cref{tab:uag-eval}, we report the benchmark suite results for the Qwen draft models used in our \gls{uag} experiments. 

\begin{table}[tb]
\centering
\caption{\textbf{\gls{sdsd} layer-pruned and fine-grained sparse draft model results on the OpenLLM Leaderboard V1 benchmarks.}}\label{tab:fine-tuned-eval}
\begin{tabular}{llcccccccc}
\toprule
Model & Variant & \makecell{Sparsity\\\%} & ARC-C & GSM8K & \makecell{Hella\\-Swag} & MMLU & \makecell{Wino\\-grande} & \makecell{TruthfulQA\\MC2} & \makecell{Mean\\ Accuracy} \\
\midrule
\multirow[c]{6}{*}{Llama--1B} & Dense & 0 & $42.58$ & $33.13$ & $59.70$ & $45.46$ & $62.19$ & $43.85$ & $47.82$ \\
\cline{2-10}
 & Layer-pruned & 50 & $25.85$ & $0.08$ & $34.74$ & $25.42$ & $52.25$ & $42.50$ & $30.14$ \\
\cline{2-10}
 & \multirow[c]{4}{*}{SparseGPT} & 50 & $37.03$ & $21.99$ & $52.29$ & $33.40$ & $58.72$ & $45.35$ & $41.46$ \\
 &  & 2:4 & $32.34$ & $11.98$ & $46.23$ & $28.36$ & $56.99$ & $44.04$ & $36.66$ \\
 &  & 66 & $30.38$ & $7.73$ & $40.94$ & $26.81$ & $55.17$ & $43.72$ & $34.12$ \\
 &  & 75 & $25.51$ & $1.59$ & $34.15$ & $26.08$ & $51.07$ & $45.19$ & $30.60$ \\
\cline{1-10} \cline{2-10}
\multirow[c]{6}{*}{Llama--3B} & Dense & 0 & $51.54$ & $64.59$ & $73.09$ & $59.71$ & $69.53$ & $49.72$ & $61.37$ \\
\cline{2-10}
 & Layer-pruned & 50 & $27.39$ & $0.38$ & $36.91$ & $25.12$ & $50.04$ & $44.22$ & $30.68$ \\
\cline{2-10}
 & \multirow[c]{4}{*}{SparseGPT} & 50 & $43.60$ & $50.19$ & $65.56$ & $52.81$ & $65.67$ & $49.32$ & $54.53$ \\
 &  & 2:4 & $42.75$ & $38.21$ & $59.93$ & $44.50$ & $62.59$ & $45.77$ & $48.96$ \\
 &  & 66 & $40.19$ & $29.72$ & $54.70$ & $39.52$ & $60.77$ & $48.12$ & $45.50$ \\
 &  & 75 & $33.45$ & $13.50$ & $45.51$ & $32.35$ & $57.46$ & $41.90$ & $37.36$ \\
\cline{1-10} \cline{2-10}
\multirow[c]{4}{*}{Qwen-0.5B} & Dense & 0 & $36.26$ & $21.53$ & $51.40$ & $46.99$ & $54.54$ & $41.95$ & $42.11$ \\
\cline{2-10}
 & Layer-pruned & 50 & $25.51$ & $0.15$ & $31.88$ & $25.54$ & $51.93$ & $46.94$ & $30.33$ \\
\cline{2-10}
 & \multirow[c]{2}{*}{SparseGPT} & 50 & $32.25$ & $5.84$ & $44.69$ & $38.33$ & $54.78$ & $43.16$ & $36.51$ \\
 &  & 2:4 & $30.03$ & $4.09$ & $39.98$ & $28.24$ & $54.22$ & $43.50$ & $33.35$ \\
\cline{1-10} \cline{2-10}
\multirow[c]{4}{*}{Qwen-1.5B} & Dense & 0 & $53.92$ & $31.31$ & $67.70$ & $60.35$ & $65.59$ & $46.61$ & $54.25$ \\
\cline{2-10}
 & Layer-pruned & 50 & $27.65$ & $0.76$ & $38.48$ & $25.71$ & $50.91$ & $43.73$ & $31.21$ \\
\cline{2-10}
 & \multirow[c]{2}{*}{SparseGPT} & 50 & $47.27$ & $42.76$ & $60.79$ & $50.51$ & $61.56$ & $47.98$ & $51.81$ \\
 &  & 2:4 & $39.59$ & $26.84$ & $54.44$ & $40.46$ & $58.25$ & $43.64$ & $43.87$ \\
\cline{1-10} \cline{2-10}
\multirow[c]{4}{*}{Qwen-3B} & Dense & 0 & $60.24$ & $10.54$ & $75.18$ & $66.37$ & $70.40$ & $58.76$ & $56.92$ \\
\cline{2-10}
 & Layer-pruned & 50 & $30.29$ & $1.06$ & $40.53$ & $23.57$ & $50.67$ & $43.29$ & $31.57$ \\
\cline{2-10}
 & \multirow[c]{2}{*}{SparseGPT} & 50 & $50.34$ & $61.11$ & $68.85$ & $58.27$ & $65.82$ & $54.05$ & $59.74$ \\
 &  & 2:4 & $46.93$ & $44.12$ & $62.83$ & $49.68$ & $65.04$ & $50.62$ & $53.20$ \\
\cline{1-10} \cline{2-10}
\bottomrule
\end{tabular}
\end{table}

\begin{table}[tb]
\centering
\caption{\textbf{One-shot pruned layer-pruned and fine-grained sparse draft model results on the OpenLLM Leaderboard V1 benchmarks.}}\label{tab:one-shot-eval}
\begin{tabular}{llcccccccc}
\toprule
Model & Variant & \makecell{Sparsity\\\%} & ARC-C & GSM8K & \makecell{Hella\\-Swag} & MMLU & \makecell{Wino\\-grande} & \makecell{TruthfulQA\\MC2} & \makecell{Mean\\ Accuracy} \\
\midrule
\multirow[c]{6}{*}{Llama-1B} & Dense & 0 & $42.58$ & $33.13$ & $59.70$ & $45.46$ & $62.19$ & $43.85$ & $47.82$ \\
\cline{2-10}
 & Layer-pruned & 50 & $26.62$ & $0.00$ & $27.21$ & $22.97$ & $48.54$ & $47.39$ & $28.79$ \\
\cline{2-10}
 & \multirow[c]{4}{*}{SparseGPT} & 50 & $34.98$ & $4.62$ & $48.50$ & $31.01$ & $59.35$ & $40.69$ & $36.53$ \\
 &  & 2:4 & $26.71$ & $1.06$ & $38.15$ & $26.15$ & $55.72$ & $41.85$ & $31.61$ \\
 &  & 66 & $24.74$ & $0.99$ & $32.93$ & $26.66$ & $52.49$ & $41.35$ & $29.86$ \\
 &  & 75 & $22.53$ & $0.30$ & $27.68$ & $25.45$ & $47.91$ & $46.84$ & $28.45$ \\
\cline{1-10} \cline{2-10}
\multirow[c]{6}{*}{Llama--3B} & Dense & 0 & $51.54$ & $64.59$ & $73.09$ & $59.71$ & $69.53$ & $49.72$ & $61.37$ \\
\cline{2-10}
 & Layer-pruned & 50 & $25.00$ & $0.00$ & $28.14$ & $24.04$ & $49.41$ & $49.77$ & $29.39$ \\
\cline{2-10}
 & \multirow[c]{4}{*}{SparseGPT} & 50 & $42.75$ & $31.99$ & $62.44$ & $49.22$ & $63.46$ & $47.66$ & $49.59$ \\
 &  & 2:4 & $33.62$ & $5.46$ & $47.78$ & $37.48$ & $58.17$ & $44.72$ & $37.87$ \\
 &  & 66 & $28.24$ & $1.44$ & $39.68$ & $30.05$ & $57.38$ & $43.90$ & $33.45$ \\
 &  & 75 & $23.98$ & $0.15$ & $29.52$ & $25.61$ & $50.67$ & $45.64$ & $29.26$ \\
\cline{1-10} \cline{2-10}
\multirow[c]{4}{*}{Qwen-0.5B} & Dense & 0 & $36.26$ & $21.53$ & $51.40$ & $46.99$ & $54.54$ & $41.95$ & $42.11$ \\
\cline{2-10}
 & Layer-pruned & 50 & $21.33$ & $0.00$ & $27.61$ & $22.98$ & $50.83$ & $49.69$ & $28.74$ \\
\cline{2-10}
 & \multirow[c]{2}{*}{SparseGPT} & 50 & $29.35$ & $3.03$ & $42.74$ & $33.04$ & $53.91$ & $42.85$ & $34.15$ \\
 &  & 2:4 & $25.09$ & $1.44$ & $34.61$ & $25.92$ & $52.57$ & $43.50$ & $30.52$ \\
\cline{1-10} \cline{2-10}
\multirow[c]{4}{*}{Qwen-1.5B} & Dense & 0 & $53.92$ & $31.31$ & $67.70$ & $60.35$ & $65.59$ & $46.61$ & $54.25$ \\
\cline{2-10}
 & Layer-pruned & 50 & $24.06$ & $0.00$ & $32.25$ & $24.37$ & $49.33$ & $48.14$ & $29.69$ \\
\cline{2-10}
 & \multirow[c]{2}{*}{SparseGPT} & 50 & $43.94$ & $14.25$ & $57.65$ & $48.60$ & $61.01$ & $42.87$ & $44.72$ \\
 &  & 2:4 & $31.23$ & $2.65$ & $43.99$ & $35.71$ & $56.04$ & $40.74$ & $35.06$ \\
\cline{1-10} \cline{2-10}
\multirow[c]{4}{*}{Qwen-3B} & Dense & 0 & $60.24$ & $10.54$ & $75.18$ & $66.37$ & $70.40$ & $58.76$ & $56.92$ \\
\cline{2-10}
 & Layer-pruned & 50 & $23.29$ & $0.08$ & $30.87$ & $24.78$ & $50.51$ & $49.39$ & $29.82$ \\
\cline{2-10}
 & \multirow[c]{2}{*}{SparseGPT} & 50 & $48.98$ & $43.44$ & $66.22$ & $55.64$ & $66.93$ & $52.77$ & $55.66$ \\
 &  & 2:4 & $39.08$ & $8.64$ & $53.25$ & $43.95$ & $60.93$ & $42.01$ & $41.31$ \\
\cline{1-10} \cline{2-10}
\bottomrule
\end{tabular}
\end{table}

\begin{table}[tb]
\centering
\caption{\textbf{OpenLLM Leaderboard V1 benchmarks for \gls{sdsd} layer-pruned and fine-grained sparse Qwen-2.5 draft models aligned with a Llama-3.1-70B-Instruct target model for the \gls{uag} setting.}}\label{tab:uag-eval}
\begin{tabular}{llcccccccc}
\toprule
Model & Variant & \makecell{Sparsity\\\%} & ARC-C & GSM8K & \makecell{Hella\\-Swag} & MMLU & \makecell{Wino\\-grande} & \makecell{TruthfulQA\\MC2} & \makecell{Mean\\ Accuracy} \\
\midrule
\multirow[c]{3}{*}{Qwen-0.5B} & Layer-pruned & 50 & $26.88$ & $0.00$ & $32.19$ & $25.35$ & $50.12$ & $45.33$ & $29.98$ \\
\cline{2-10}
 & \multirow[c]{2}{*}{SparseGPT} & 50 & $32.08$ & $6.29$ & $44.83$ & $38.07$ & $55.88$ & $44.56$ & $36.95$ \\
 &  & 2:4 & $28.84$ & $4.55$ & $40.50$ & $28.00$ & $54.70$ & $44.34$ & $33.49$ \\
\cline{1-10} \cline{2-10}
\multirow[c]{3}{*}{Qwen-1.5B} & Layer-pruned & 50 & $28.75$ & $0.08$ & $39.21$ & $25.57$ & $52.25$ & $43.67$ & $31.59$ \\
\cline{2-10}
 & \multirow[c]{2}{*}{SparseGPT} & 50 & $45.48$ & $42.53$ & $60.32$ & $51.30$ & $61.33$ & $46.25$ & $51.20$ \\
 &  & 2:4 & $38.65$ & $28.20$ & $53.92$ & $40.09$ & $58.64$ & $43.76$ & $43.88$ \\
\cline{1-10} \cline{2-10}
\multirow[c]{3}{*}{Qwen-3B} & Layer-pruned & 50 & $31.48$ & $0.99$ & $40.55$ & $22.98$ & $50.28$ & $42.17$ & $31.41$ \\
\cline{2-10}
 & \multirow[c]{2}{*}{SparseGPT} & 50 & $48.98$ & $58.07$ & $67.88$ & $57.61$ & $65.04$ & $49.09$ & $57.78$ \\
 &  & 2:4 & $44.71$ & $45.94$ & $61.77$ & $48.72$ & $63.93$ & $48.03$ & $52.18$ \\
\cline{1-10} \cline{2-10}
\bottomrule
\end{tabular}
\end{table}

\FloatBarrier

\end{document}

%% file: math_commands.tex
\usepackage{amsmath,amsfonts,bm}

\def\eqref#1{equation~\ref{#1}}

\def\1{\bm{1}}

\DeclareMathAlphabet{\mathsfit}{\encodingdefault}{\sfdefault}{m}{sl}
\SetMathAlphabet{\mathsfit}{bold}{\encodingdefault}{\sfdefault}{bx}{n}

\DeclareMathOperator*{\argmax}{arg\,max}
\DeclareMathOperator*{\argmin}{arg\,min}

%% file: glossary.tex
\newacronym{dnn}{DNN}{Deep Neural Network}
\newacronym{cnn}{CNN}{Convolutional Neural Network}
\newacronym{nn}{NN}{Neural Network}
\newacronym{ann}{ANN}{Artificial Neural Network}
\newacronym{snn}{SNN}{Sparse Neural Network}
\newacronym{lt}{LT}{Lottery Ticket}
\newacronym{ntk}{NTK}{the Neural Tangent Kernel}
\newacronym{lth}{LTH}{Lottery Ticket Hypothesis}
\newacronym{nlp}{NLP}{Natural Language Processing}
\newacronym{dst}{DST}{Dynamic Sparse Training}
\newacronym{gpu}{GPU}{Graphics Processing Unit}
\newacronym{pi}{PI}{Primary Investigator}
\newacronym{ai}{AI}{Artificial Intelligence}
\newacronym{ml}{ML}{Machine Learning}
\newacronym{cv}{CV}{Computer Vision}
\newacronym{neurips}{NeurIPS}{Neural Information Processing Systems}
\newacronym{iclr}{ICLR}{the International Conference on Learning Representations}
\newacronym{cvpr}{CVPR}{the IEEE/CVF Conference on Computer Vision and Pattern Recognition}
\newacronym{flops}{FLOPs}{Floating Point Operations}
\newacronym{relu}{ReLU}{Rectified Linear Unit}
\newacronym{srigl}{SRigL}{Structured RigL}
\newacronym{rigl}{RigL}{Rigging the Lottery Ticket}
\newacronym{set}{SET}{Sparse Evolutionary Training}
\newacronym{erk}{ERK}{Erd\H{o}s-R\'enyi-Kernel}
\newacronym{ste}{STE}{Straight-Through Estimator}
\newacronym{srste}{SR-STE}{Sparse-Refined Straight-Through Estimator}
\newacronym{ilsvrc}{ILSVRC-12}{2012 ImageNet Large Scale Visual Recognition Challenge}
\newacronym{ast}{AST}{Alternating Sparse Training}
\newacronym{deepr}{DeepR}{Deep Rewiring}
\newacronym{snfs}{SNFS}{Sparse Networks from Scratch}
\newacronym{dsr}{DSR}{Dynamic Sparse Reparameterization}
\newacronym{topkast}{Top-KAST}{Top-K Always Sparse Training}
\newacronym{mest}{MEST}{Memory-Economic Sparse Training}
\newacronym{dsb}{DSB}{Dynamic Shuffled Block}
\newacronym{dynsparse}{DynSparse}{Dynamic Sparsity}
\newacronym{vit}{ViT}{Vision Transformer}
\newacronym{cpu}{CPU}{Central Processing Unit}
\newacronym{csr}{CSR}{Compressed Sparse Row}
\newacronym{saow}{SAOW}{Sparse Adapters with Outlier Weights}
\newacronym{llm}{LLM}{Large Language Model}
\newacronym{rag}{RAG}{Retrieval Augmented Generation}
\newacronym{mlp}{MLP}{Multi-Layer Perceptron}
\newacronym{ees}{EES}{Exploiting Emergent Structure}
\newacronym{drl}{DRL}{Deep Reinforcement Learning}
\newacronym{hpc}{HPC}{High Performance Computing}
\newacronym{ddp}{DDP}{Distributed Data Parallel}
\newacronym{hqp}{HQP}{Highly Qualified Personnel}
\newacronym{moe}{MoE}{Mixture-of-Experts}
\newacronym{nas}{NAS}{Neural Architecture Search}
\newacronym{tpu}{TPU}{Tensor Processing Unit}
\newacronym{dram}{DRAM}{Dynamic Random Access Memory}
\newacronym{simd}{SIMD}{Single Instruction Multiple Data}
\newacronym{peft}{PEFT}{Parameter Efficient Fine-Tuning}
\newacronym{qat}{QAT}{Quantization-Aware Training}
\newacronym{ptq}{PTQ}{Post-Training Quantization}
\newacronym{sanr}{SANR}{Sparsity-Aware Normalization and Regularization}
\newacronym{squash}{SQuAsh}{Sparse, Quantized fine-tuning using Adapters at home}
\newacronym{drac}{DRAC}{Digital Research Alliance of Canada}
\newacronym{rac}{RAC}{Resource Allocation Competition}
\newacronym{kd}{KD}{Knowledge Distillation}
\newacronym{hbm}{HBM}{High-Bandwidth Memory}
\newacronym{sdsd}{SD$^2$}{Self-Distilled Sparse Drafters}
\newacronym{ar}{AR}{Acceptance Rate}
\newacronym{mac}{MAC}{Multiply-Accumulate operation}
\newacronym{uag}{UAG}{Universal Assisted Generation}
\newacronym{owl}{OWL}{Outlier Weighted Layer-wise sparsity}
\newacronym{mal}{MAL}{Mean Accepted Length}
\newacronym{lm-head}{LM head}{language modelling head}